%% file: drops_to_grid.tex
\documentclass[10pt,twocolumn,letterpaper]{article}

\usepackage{cvpr}              

\input{preamble}

\definecolor{cvprblue}{rgb}{0.21,0.49,0.74}
\usepackage[pagebackref,breaklinks,colorlinks,allcolors=cvprblue]{hyperref}

\usepackage[accsupp]{axessibility}  

\def\paperID{41994} 
\def\confName{CVPR}
\def\confYear{2026}

\title{From Drops to Grid: Noise-Aware Spatio-Temporal \\ Neural Process for Rainfall Estimation}

\author{%
  Rafael Pablos Sarabia \\
  Dept. Comp. Sc., Aarhus\\ University \& Cordulus \\
  {\tt\small rpablos@cs.au.dk} \\
  \and
  Joachim Nyborg \\
  Cordulus \\
  {\tt\small jn@cordulus.com} \\
  \and
  Morten Birk \\
  Cordulus \\
  {\tt\small mb@cordulus.com} \\
  \and
  Ira Assent \\
  Dept. of Comp. Science\\
  Aarhus University \\
  {\tt\small ira@cs.au.dk} \\
}

\begin{document}
\maketitle
\input{sections/0_abstract}

\input{sections/1_intro}
\input{sections/2_related_work}
\input{sections/3_method}
\input{sections/4_experiments}
\input{sections/5_conclusion}

\clearpage

\section*{Acknowledgments}
This work is partly funded by the Innovation Fund Denmark (IFD) under File No. 2052-00064B, as well as by Danish Pioneer Centre for AI\footnote{\url{https://aicentre.dk}}, DNRF grant number P1.

{
    \small
    \bibliographystyle{ieeenat_fullname}
    \bibliography{main}
}

\clearpage
\setcounter{page}{1}
\maketitlesupplementary
\input{sections/appendix/01_zig}
\input{sections/appendix/02_training}
\input{sections/appendix/03_metrics}
\input{sections/appendix/04_baselines}
\input{sections/appendix/05_visualizations}
\input{sections/appendix/06_dl_baselines}
\input{sections/appendix/07_ablation}
\input{sections/appendix/08_stations}
\input{sections/appendix/09_eu}

\end{document}

%% file: preamble.tex
\usepackage{multirow}
\usepackage{stfloats}
\usepackage{bbm}
\usepackage{graphicx}

%% file: sections/0_abstract.tex
\begin{abstract}
High-resolution rainfall observations are crucial for weather forecasting, water management, and hazard mitigation. Traditional operational measurements are often biased and low-resolution, limiting their ability to capture local rainfall. Accurate high-resolution rainfall maps require integrating sparse surface observations, yet existing deep learning densification methods are hindered by rainfall's skewed, localized nature, noise, and limited spatio-temporal fusion. We present \mbox{DropsToGrid}, a Neural Process–based method that generates dense rainfall fields by fusing temporal sequences from noisy, irregularly distributed private weather stations with spatial context from radar. Leveraging multi-scale feature extraction, temporal attention, and multi-modal fusion, the model produces stochastic, continuous rainfall estimates and explicitly quantifies uncertainty. Evaluations on real-world datasets demonstrate that DropsToGrid outperforms both operational and deep learning baselines, generating accurate high-resolution rainfall maps with well-calibrated uncertainty, even when only few stations are available and in cross-regional scenarios. Code is available at \small{\url{https://github.com/rafapablos/DropsToGrid}}
\end{abstract}

%% file: sections/1_intro.tex
\section{Introduction}
\label{sec:intro}

Rainfall drives the movement of water from the atmosphere to the Earth's surface, influencing ecosystems, agriculture, infrastructure, and disaster management through events like rain, droughts, or floods~\cite{precipitation_importance}. Accurate and timely rainfall observations are crucial for weather and climate prediction, water management, and hazard forecasting. Capturing rainfall involves estimating a continuous gridded field of rainfall intensity over space and time by integrating heterogeneous observations from gauges, radar, and satellites \cite{opera_1, opera_2, imerg, climategrid, era5}. However, generating rain grids at high spatial and temporal resolution remains challenging due to the local variability and intermittent nature of rainfall~\cite{precipitation_nature}.

\begin{figure}[t]
  \centering
   \includegraphics[width=0.84\linewidth]{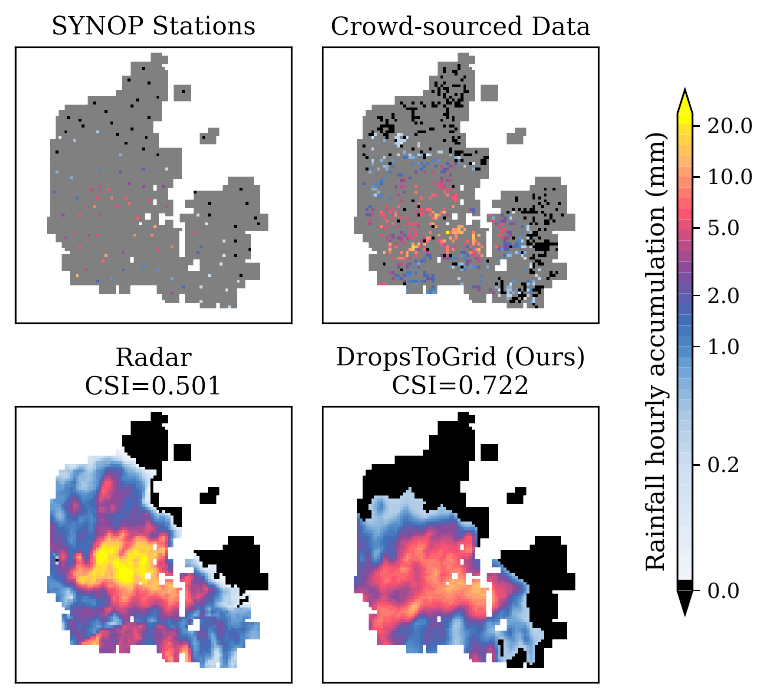}
   \caption{Precipitation estimates. Top: rainfall data from SYNOP stations show only sparse coverage (left); crowd-sourced PWS observations are unevenly distributed and typically suffer from quality issues (right). Bottom: radar-based estimates exhibit substantial biases (left); our proposed DropsToGrid method produces reliable, high-quality dense precipitation maps from sparse PWS data and radar (right). Gray: missing data; white: outside scope.}
   \label{fig:motivation}
\end{figure}

Ground-based rain gauges provide the most direct and physically accurate surface rainfall measurements~\cite{radar_qpe}. National SYNOP (surface synoptic observations) networks, designed for synoptic scales (100-3,000km), are often too sparse to capture microscale ($<$100m) or local-scale (100m-3km) variability~\cite{wmo_stations} (cf. Figure~\ref{fig:motivation} top left). Remote sensing products offer dense coverage but have limitations: radar (Figure~\ref{fig:motivation} bottom left) measures reflectivity, which requires bias-prone empirical conversion to rainfall~\cite{reflectivity}; satellite estimates provide global but unreliable coverage over regions (e.g. Europe~\cite{precipitation_comparison}); reanalysis integrates multiple sources but suffers from low resolution, model assumptions, delays, and biases in surface rainfall~\cite{era5-eval}. Thus, none of these approaches can accurately capture localized rainfall dynamics.

Privately-owned weather stations (PWS) represent a growing source of observations \cite{pws_growth}. Crowd-sourced PWS networks often exceed traditional SYNOP coverage but vary widely in hardware quality, installation, and maintenance, resulting in noisy, biased, and sometimes inconsistent measurements \cite{pws_errors} (Figure~\ref{fig:motivation}). Moreover, these networks are not static: stations may appear, disappear, or move over time, producing a constantly changing and spatially irregular observation pattern. This creates a challenging learning problem: models must infer gridded, high-resolution rainfall fields from irregular, sparse, and noisy inputs. Unlike standard image reconstruction, the available "pixels" are irregular in space, evolve over time, and are too noisy to use directly as outputs. This requires spatio-temporal reasoning, uncertainty-aware processing, and multi-modal fusion.

Our goal is to learn to densify rainfall fields by jointly leveraging sparse PWS measurements and radar observations, making the task both a spatial densification and multi-modal fusion problem. From a learning perspective, this requires models that can infer dense spatio-temporal rainfall patterns from irregular, noisy time series while using radar to provide rich spatial context. Radar and station measurements capture complementary but different physical quantities: radar reflects atmospheric moisture, whereas stations measure surface rainfall. As the mapping between radar intensity and surface rainfall is complex, non-local, and context-dependent \cite{station_radar}, naive pixel-level alignment can be misleading. A robust approach must learn flexible, non-linear cross-modal mappings and explicitly account for uncertainty and sensor-specific bias to capture complex, context-dependent relationships and avoid propagating errors when densifying the rainfall field.

We propose DropsToGrid, a Neural Process–based method for rainfall densification that integrates temporal station sequences with radar-derived spatial context. Neural Processes (NPs)~\cite{cnp, npf} are stochastic models handling variable-size context sets and providing uncertainty-aware inference without retraining for new input configurations. DropsToGrid leverages NPs to learn a stochastic, continuous rainfall representation from sparse, noisy PWS observations, guided by dense radar context (Fig.~\ref{fig:motivation}). It captures local and large-scale rainfall patterns via multi-scale feature extraction, temporal attention over station sequences, multi-modal attention, and translation-equivariant spatial fusion. By explicitly modeling uncertainty, DropsToGrid produces calibrated predictions even in poorly observed regions.

We evaluate DropsToGrid on real-world rainfall datasets against operational estimators and deep learning baselines using NPs for station densification. Our analysis includes ablation studies, the effect of station density on performance, assessments of spatial uncertainty, and cross-regional Europe-wide evaluation. Results show that DropsToGrid generates accurate high-resolution rainfall estimates from sparse, irregular, and noisy data, along with well-calibrated uncertainty maps.

To our knowledge, DropsToGrid is the first Neural Process‑based approach for densifying rainfall from PWS data, combining temporal sequences and radar context in a single probabilistic method. Our main contributions are:
\begin{itemize}
    \item A Neural Process-based method for probabilistic rainfall densification exploiting temporal PWS sequences and radar context.
    \item A multi-modal attention mechanism with translation-equivariant fusion for spatial and temporal dependencies.
    \item Thorough empirical study with operational and deep learning baselines, uncertainty and station density analysis, cross-regional Europe-wide evaluation, and ablations.
\end{itemize}

%% file: sections/2_related_work.tex
\section{Related work}
\label{sec:related_work}

\textbf{Rainfall estimation and forecasting models.} 
Recent progress in medium-range weather forecasting is driven by e.g. GraphCast~\cite{graphcast}, Pangu-Weather~\cite{panguweather}, or GenCast~\cite{gencast} models of reanalysis data like ERA5~\cite{era5}. Reanalysis combines historical observations with numerical weather models to produce consistent, gridded estimates of atmospheric variables over time. However, ERA5 suffers from systematic biases, coarse spatial resolution, temporal delays, and limited use of direct observations, especially for surface variables and rainfall~\cite{era5-eval}. Rainfall remains one of the hardest variables to predict and is often underrepresented in such models~\cite{weatherbench2}. NeuralGCM~\cite{neuralgcm-precipitation} incorporates satellite observations for global rainfall but lacks the spatial resolution and update rate required for high-frequency applications.

Radar-based nowcasting dominates short-term rainfall modeling~\cite{sevir, meteonet, hko7}, with models like Earthformer~\cite{earthformer}, CasCast~\cite{cascast}, and DiffCast~\cite{diffcast} achieving highest accuracy in dense regions. However, radar rainfall often diverges from surface observations due to attenuation, beam blocking, range dependence, and other biases~\cite{radar_qpe, radar_issues}. 
To better capture surface-level observations, recent work integrates station data. MetNet-3~\cite{metnet3} uses stations to improve temperature, dew point, and wind forecasts, but rainfall remains difficult due to sparsity and irregularity. End-to-end frameworks such as Aardvark~\cite{aardvark}, FengWu-4DVar~\cite{fengwu4d}, FuXi-DA~\cite{fuxida}, and ECMWF's ongoing observation integration efforts~\cite{ecmwf_obs} show promise in combining unstructured observations to improve forecasts and reduce computational cost~\cite{gridded-tnp}. Still, most exclude rainfall or limit it to coarse accumulations, e.g., six-hour totals in FuXi-DA~\cite{fuxida}. 

\textbf{Weather stations for rainfall estimation.} 
Estimating rainfall from sparse, noisy stations is challenging. Classical spatial interpolation methods such as Inverse Distance Weighting (IDW), Kriging, and Gaussian Processes (GPs) trade off simplicity, accuracy, and computational cost.
IDW~\cite{idw} computes distance-weighted averages, assuming smoothly varying fields, but neglects spatial autocorrelation and underperforms on complex terrain or non-uniform rainfall. Kriging~\cite{kriging} models spatial covariance and yields uncertainty estimates via kriging variance~\cite{kriging_1, kriging_2, kriging_3, kriging, kriging_4, kriging_5}. The latter is generally superior but computationally costly with poor scalability~\cite{rainfall_kriging}. GPs~\cite{gp_ml} generalize Kriging within a Bayesian framework, providing principled uncertainty quantification but scaling cubically with the number of observations. 
Performance of these models degrades with irregular networks, complex terrain, and intermittent rainfall, and their computational cost hinders dense spatio-temporal predictions \cite{idw_issue_clusters, idw_terrain_issue, rainfall_kriging, npf, ccnp-downscaling}. These issues motivate data-driven models that can learn flexible spatio-temporal representations and leverage multi-modal data (e.g., radar, satellite) for accuracy and scalability.

\textbf{Neural Processes for densification.} 
Neural Processes~\cite{cnp, npf} offer uncertainty modeling of Gaussian Processes and scalability of neural networks, enabling efficient, uncertainty-aware inference from irregular observations.
Conditional Neural Process (CNP)~\cite{cnp} is an encoder–decoder framework mapping context to target points but often underfits due to global latent representation. Attentive CNPs (AttnCNPs)~\cite{anp, attention_translation} use attention to weigh context relevance, improving local fidelity at the cost of higher dependence on training coverage. Convolutional CNPs (ConvCNPs)~\cite{ccnp} add translation equivariance, improving local sensitivity and extrapolation - ideal for geospatial and meteorological tasks. Extensions include Relational CNPs (RCNPs)~\cite{rcnp}, Group-Equivariant CNPs~\cite{equivcnp}, and Steerable-Equivariant CNPs~\cite{steercnp}.

Recent transformer-based variants~\cite{tnp, tetnp, gridded-tnp}) jointly model dense gridded and sparse stations. However, most NP-based climate applications focus on smoother variables, such as temperature~\cite{ccnp-downscaling, gridded-tnp} or coarse daily rainfall~\cite{ccnp-downscaling}, often in synthetic or noise-free settings~\cite{gridded-tnp}. Rainfall densification remains largely unexplored under real-world conditions with noisy, irregular, and heterogeneous observations, and we compare against current approaches \cite{gridded-tnp} in Sec.~\ref{sec:experiments}.

%% file: sections/3_method.tex
\section{Method}
\label{sec:method}

High-resolution rainfall estimation is challenging due to the unreliable and heterogenous nature of available observations. In this work, we focus on integrating sparse, noisy, and irregularly distributed private weather station (PWS) measurements with radar data that, while spatially continuous, has biases and errors. Because rain gauges provide the most accurate surface rainfall measurements, they serve as the reference for both targets and evaluation. Our goal is to learn a probabilistic mapping from these complementary sources to dense, uncertainty-aware rainfall grids. Crucially, DropsToGrid must be translation equivariant: spatial shifts in the input should produce equivalent shifts in the output. This ensures consistent predictions as rainfall systems move and enables generalization across regions without retraining for specific station layouts or fixed grids.

We introduce DropsToGrid, a Neural Process method for probabilistic rainfall densification that fuses temporally evolving PWS measurements with radar observations. DropsToGrid (i) represents rainfall as a stochastic spatial field, (ii) is robust to missing or noisy inputs, and (iii) produces calibrated uncertainty over dense grids. It consists of two main components:
\begin{enumerate}
    \item \textbf{A translation-equivariant spatio-temporal neural architecture} that encodes stations and radar with SetConv layers, processes them in a U-Net encoder-decoder, and fuses modalities and time with an attention bottleneck.
    \item \textbf{A noise-aware Zero-Inflated Gamma (ZIG) objective} modeling rainfall's sparsity and skewed distribution.
\end{enumerate}

\begin{figure*}[!t]
  \centering
   \includegraphics[width=\linewidth]{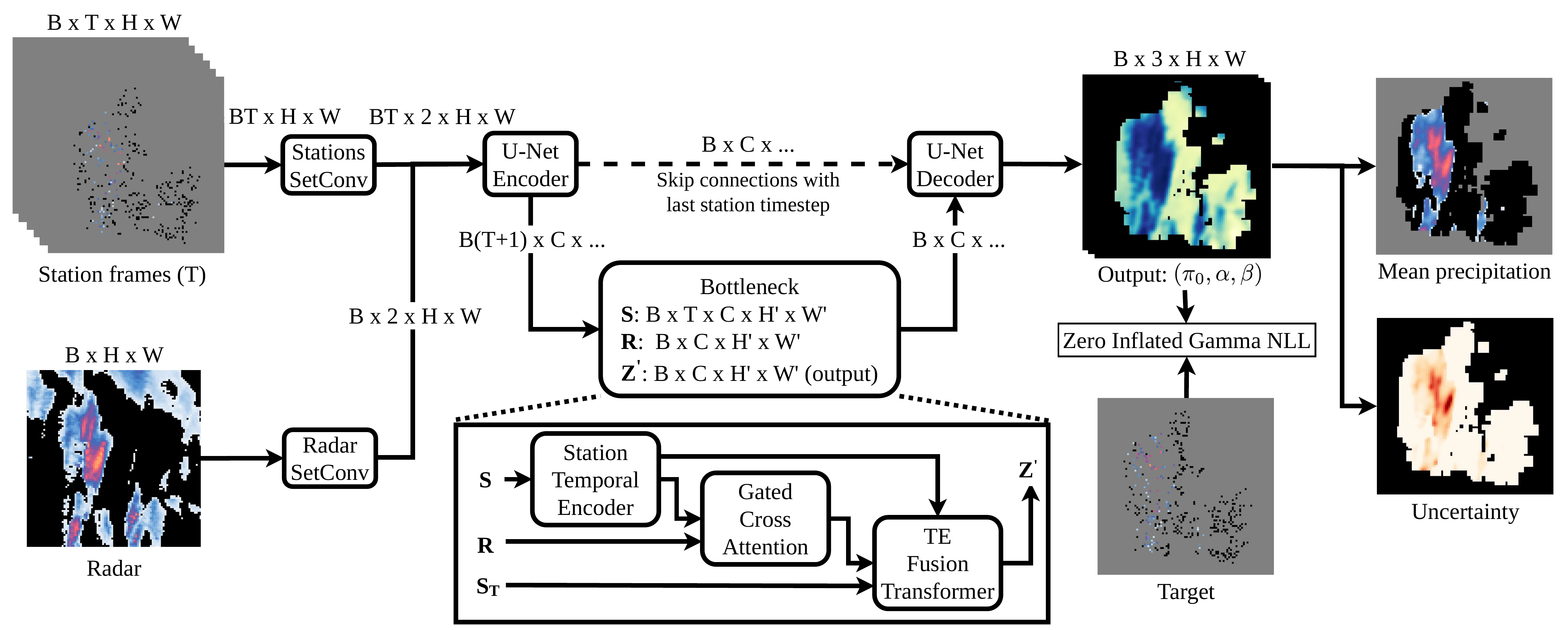}
   \caption{Overview of the DropsToGrid architecture. Sparse, temporally evolving rainfall measurements from private weather stations (PWS) and spatially continuous radar observations are encoded via a SetConv and U-Net architecture. A spatio-temporal bottleneck fuses radar and temporal context, and a probabilistic decoder outputs dense rainfall distributions, producing both mean estimates and calibrated uncertainty. Attention is applied at the latent resolution to capture long-range dependencies efficiently.}
   \label{fig:arch}
\end{figure*}

\subsection{DropsToGrid architecture}
DropsToGrid builds on Convolutional Conditional Neural Process (ConvCNP)~\cite{ccnp}, enabling meta-learning of a stochastic process or distribution over predictors. We adopt ConvCNP for rainfall densification because its convolutional structure efficiently aggregates information from spatial neighbors, capturing local correlations that dominate rainfall patterns. Given a set of observed \textit{context} points $\mathcal{C} = \{(x^{(c)}, y^{(c)})\}_{c=1}^N$, where $x^{(c)}=(u^{(c)}, v^{(c)}, t^{(c)})$ are spatio-temporal coordinates and $y^{(c)}$ the rainfall intensity from PWS stations, the goal is to model the conditional distribution $p_{\theta}(\mathbf{y}_{\mathcal{T}}| \mathbf{x}_{\mathcal{T}}; \mathcal{C})$ for \textit{target} locations $\mathcal{T} = \{{x^{(i)}}\}_{i=1}^I$, with $\theta$ denoting network parameters. Following a meta-learning approach, DropsToGrid is trained across multiple spatio-temporal episodes to rapidly adapt to new contexts and infer predictive distributions for unseen targets~\cite{npf}.

DropsToGrid extends ConvCNP to handle heterogeneous, non–pixel-aligned sources with temporally evolving data. Sparse PWS measurements are encoded via Set Convolution (SetConv)~\cite{ccnp} layers, mapped to high-dimensional latent space through a U-Net-style encoder~\cite{unet}. This representation is refined via a temporal, cross-modal, and spatial bottleneck that fuses temporal context and radar data. A probabilistic decoder then outputs rainfall distributions for the target time, producing mean estimates and calibrated uncertainty maps (Fig.~\ref{fig:arch}). DropsToGrid captures rainfall features across scales, from local convective cells to large synoptic systems in an efficient manner. Convolutions extract local spatial dependencies, and attention mechanisms operate in compact latent spaces at lowest resolution in the U-Net, balancing expressiveness and computational cost. This design preserves translation equivaraince, ensuring consistent behavior as rainfall fields move across space and enabling generalization without retraining. 

\subsubsection{Sparse-to-dense encoding}
Rainfall measurements from stations are often irregularly spaced: dense in some regions, sparse in others, and occasionally missing entirely, while radar observations, though gridded, can have varying reliability due to beam blockage or attenuation. Standard convolutions, which assume regular grids, struggle with these irregularities, leading to artifacts or loss of fine-grained detail. To address this, we project the context set into a continuous latent field using SetConv, which is permutation-invariant and supports variable-sized inputs. Each observation, defined by spatial coordinates and intensity, is encoded as a localized contribution and convolved over nearby points to form a dense latent grid, effectively translating sparse inputs into a smooth, continuous representation for rainfall densification.

Overlapping observations pose another challenge: naive aggregation inflates signals in dense regions while underrepresenting sparse areas. SetConv incorporates a density channel to normalize overlapping contributions. Each observation contributes a Gaussian-shaped bump with learnable width, ensuring smooth spatial aggregation. The sparse station observations are projected onto a spatial grid to form the input $I$, while the corresponding mask $M$ encodes whether each grid cell contains a valid measurement or a placeholder value. For gridded input $I$ with mask $M$:
\begin{equation*}
\text{signal} = \mathrm{Conv}_\text{abs}(I \odot M), \quad
\text{density} = \mathrm{Conv}_\text{abs}(M)
\end{equation*}
\begin{equation*}
\mathrm{SetConv}(I, M) = \left[ \frac{\text{signal}}{\text{density} + \epsilon}, \, \text{density} \right]
\end{equation*}
where $\mathrm{Conv}_\text{abs}$ is a depthwise convolution with non-negative weights, $\odot$ denotes element-wise multiplication, and $\epsilon$ ensures numerical stability.

Because station and radar inputs differ in sampling density and in spatial structure, one shared encoder could dominate or distort the other in aggregation. To prevent this interference, DropsToGrid employs two independent SetConv encoders, one for stations and one for radar, allowing each to aggregate local information according to its own characteristics. Each encoder outputs two channels per timestep (normalized signal and density), which are concatenated to produce dense latent representations $\mathbf{x}^{(\text{stations})}$, $\mathbf{x}^{(\text{radar})}$. 

These dense latent representations are then embedded using modality-specific Multi-Layer Perceptrons (MLPs) to produce richer feature vectors for each pixel. For each modality $m \in \{\text{station}, \text{radar}\}$:
\begin{equation*}
\mathbf{z}^{(m)}_{b,t,h,w} = f^{(m)}_{\mathrm{MLP}}\left( \mathbf{x}^{(m)}_{b,t,h,w} \right),
\end{equation*}
where $f^{(m)}_{\mathrm{MLP}}$ is a pointwise feed-forward network and $(b, t, h, w)$ index batch, time, height, and width.

\subsubsection{Spatial feature extraction with bottleneck fusion}
Embeddings $\mathbf{Z}^{(\text{stations})}, \quad \mathbf{Z}^{(\text{radar})} \in \mathbb{R}^{B \times T \times C \times H \times W}$ capture rich, dense, pixel-wise information per source and timestep. These embeddings cannot be merged along the channel dimension, since radar and stations measure complementary but different physical quantities (moisture vs rainfall) with complex, non-local, and context-dependent relationships \cite{station_radar}. To address this, a 2D U-Net processes each temporal slice and source independently, stacking them along the batch axis. Time is kept separate so the bottleneck can model temporal dependencies, capturing short- and long-term patterns from noisy input sequences while extracting spatial features per source and time step. At the U-Net bottleneck, we obtain compact feature maps $\mathbf{S} \in \mathbb{R}^{B \times T \times C \times H' \times W'}$ (stations) and $\mathbf{R} \in \mathbb{R}^{B \times C \times H' \times W'}$ (radar). These are fused through a three-part block modeling temporal, cross-modal, and spatial relationships to produce a latent representation that integrates both sources.
\paragraph{Station temporal encoder.}
Temporal dependencies across the $T$ station timesteps are captured by a Temporal Pixelwise Transformer with a learned global query \(\mathbf{Q}_{\text{learned}} \in \mathbb{R}^{d_h}\) shared across all pixels, where \(d_h\) is the head dimensionality. This transformer operates independently on each pixel, allowing it to model short- and long-range temporal patterns for each spatial location. The global learned query acts like a summary token, aggregating information from all timesteps per pixel. Since transformers are permutation invariant, we introduce temporal order via learned positional embeddings, which flexibly encode relative importance and order of timesteps:
\begin{equation*}
    \mathbf{S}_t \leftarrow \mathbf{S}_t + \mathbf{P}_t, \quad t = 1, \dots, T,
\end{equation*}
where \(\mathbf{P}_t \in \mathbb{R}^{C}\) is a learnable vector for each timestep.

We obtain keys and values from temporally independent features, and the query from the learned global query:
\begin{equation*}
    \mathbf{K} = \mathbf{S} \mathbf{W}_K, \quad
    \mathbf{V} = \mathbf{S} \mathbf{W}_V, \quad
    \mathbf{Q} = \mathbf{Q}_{\text{learned}} \mathbf{W}_Q.
\end{equation*}

This attention yields a temporal summary for each pixel:
\begin{equation*}
\mathbf{S}_{\text{summary}} = 
\text{softmax}\left(\frac{\mathbf{QK}^\top}{\sqrt{d_h}}\right) \mathbf{V},
\end{equation*}
with $\mathbf{S}_{\text{summary}} \in \mathbb{R}^{B \times C \times H' \times W'}$. A single learned query aggregates all timesteps, yielding a compact per-pixel representation that captures  short- and long-term temporal dependencies. Finally, a transformer block applies a position-wise feed-forward network, enhancing its expressive power. 
\paragraph{Gated cross-attention.}
We adopt a Translation-Equi\-variant (TE) transformer to enable cross-modal spatial reasoning between station- and radar-based representations. The transformer performs cross-attention where the temporally fused station features act as queries, and the radar features serve as keys and values. This design allows each ground observation to selectively attend to spatial regions in the radar domain most relevant to its context, ensuring that sparse station measurements are effectively aligned with the dense but potentially contradicting radar data.

Given station features $\mathbf{S}_{\text{summary}}$ and radar features $\mathbf{R}$, the transformer first projects them into shared embedding spaces via learned linear projections and generates the output $\hat{\mathbf{R}}\in [0,1]^{B \times C \times H' \times W'}$ considering feature similarity and relative distances with the TE-Transformer:
\begin{equation*}
    \mathbf{Q} = \mathbf{S_\text{summary}}\mathbf{W}_Q, \quad
    \mathbf{K} = \mathbf{R}\mathbf{W}_K, \quad
    \mathbf{V} = \mathbf{R}\mathbf{W}_V
\end{equation*}
A learned spatial gate $G \in [0,1]^{B \times 1 \times H' \times W'}$ modulates radar contributions:
\begin{equation*}
\mathbf{R}_{\text{corrected}} = \hat{\mathbf{R}} \odot G,
\end{equation*}
where $\odot$ denotes element-wise multiplication. This gate acts as a spatial filter, emphasizing radar regions that align with station-driven attention, while suppressing spatially misaligned or uninformative regions.
\paragraph{Translation-equivariant fusion transformer.} 
Our goal is to fuse multiple feature sources while being translation-equivariant (TE) and scale-invariant, so that spatial shifts or resizing of the input lead to corresponding shifts in the output. This property is crucial for geographic generalization and preserving spatial coherence across scales. 

To achieve this, the final TE-Transformer processes a merged representation of $\mathbf{S}_{\text{summary}}$, $\mathbf{R}_{\text{corrected}}$, and the last station features $\mathbf{S}_T$, enforcing spatial translation equivariance. In TE-Transformer, instead of absolute positions, attention depends on relative displacements:
\begin{equation*}
    \Delta_{ij} = \frac{(x_i - x_j,\, y_i - y_j)}{\max(H', W')},
\end{equation*}
ensuring scale invariance. 
For each pair $(i,j)$, a learnable MLP produces pairwise attention weights based on feature similarity and geometric displacement:
\begin{equation*}
    \alpha_{ij} = f_{\text{MLP}}\big([\Delta_{ij},\, \mathbf{q}_i^\top \mathbf{k}_j]\big),
\end{equation*}
where $f_{\text{MLP}}$ is a lightweight pairwise MLP that outputs one attention logit per head. These logits are then normalized with a softmax and used to aggregate values:
\begin{equation*}
    \mathbf{z}'_i = \sum_j \text{softmax}_j(\alpha_{ij})\, \mathbf{v}_j.
\end{equation*}

By design, this ensures that spatial translation of the inputs results in equivalent translation of the output, achieving translation-equivariance and robustness to scale changes.

\subsubsection{Dense output predictions}

The bottleneck features $\mathbf{Z}' \in \mathbb{R}^{B \times C \times H' \times W'}$ are upsampled to full spatial resolution via the U-Net decoder. Unlike the encoder, which handles multiple timesteps and modalities, the decoder operates only on the last station timestep, using its corresponding skip connections to recover fine-grained spatial details essential in high-resolution rainfall fields.
A separate MLP projects decoder outputs to the required number of channels for probabilistic output parametrization.

\subsection{Noise-aware loss for rainfall}
\label{ssec:loss}
Rainfall observations exhibit heavy-tailed, zero-inflated distributions that standard regression losses or Gaussian distributions fail to capture. To model this behavior, we adopt a Zero-Inflated Gamma (ZIG)~\cite{zig-phd} likelihood as the output distribution of DropsToGrid. 
The ZIG distribution models non-negative variables with excess zeros, meaning variables that can only take values greater than or equal to zero but frequently take the value zero. In the case of accumulated rainfall, many grid cells or time steps record no rain at all, while the nonzero values can vary widely and are strongly right-skewed. To capture this pattern, the ZIG combines two components: a Bernoulli distribution that models whether rainfall occurs (zero vs. nonzero), and a Gamma distribution that models the positive rainfall intensities conditional on rain occurring. For each grid cell, DropsToGrid predicts parameters $(\pi_0, \alpha, \beta)$, where $\pi_0$ is the zero-rainfall probability, and $\alpha$ and $\beta$ are the Gamma shape and rate.
\begin{equation*}
p(y \mid \pi_0, \alpha, \beta) =
\begin{cases}
\pi_0, & y = 0, \\[6pt]
(1 - \pi_0)\,\mathrm{Gamma}(y; \alpha, \beta), & y > 0,
\end{cases}
\label{eq:zig_distribution}
\end{equation*}
\begin{equation*}
\mathrm{Gamma}(y; \alpha, \beta) =
\dfrac{\beta^{\alpha}}{\Gamma(\alpha)}\, y^{\alpha - 1} e^{-\beta y},
\quad y > 0,
\label{eq:gamma_density}
\end{equation*}
where $\Gamma(\cdot)$ is the Gamma function 
$\Gamma(\alpha) = \int_0^{\infty} t^{\alpha - 1} e^{-t}\, dt$.

In order for DropsToGrid to capture both dry and wet regimes in a unified probabilistic framework, this formulation explicitly separates the no-rain probability $\pi_0$ from the continuous rain intensities. When $\pi_0$ is large (e.g., $\pi_0>0.5$), most of the probability mass is concentrated at zero, reflecting dry conditions, whereas when $\pi_0$ is small, the mass shifts to the Gamma component, representing wet conditions with continuous rainfall intensities.
%
%
Directly training on all available station measurements can lead the model to overfit to noise or learn trivial input-output mappings, especially since rainfall observations are noisy and zero-inflated. To mitigate this, we maximize the ZIG log-likelihood using only a carefully selected set of target grid cells. Specifically, the negative log-likelihood loss is:
\begin{equation*}
\mathcal{L}_{\text{ZIG}} =
- \sum_{x \in \mathcal{T}} \log p(y(x) \mid \pi_0(x), \alpha(x), \beta(x)),
\label{eq:zig_loss}
\end{equation*}
where $\mathcal{T}$ denotes the set of grid cells with available station data that are not part of the holdout test split. Inputs are excluded from $\mathcal{T}$ to prevent DropsToGrid from memorizing noisy measurements and encourage generalization to unseen locations.
The log-likelihood for each sample is:
\begin{equation*}
\begin{aligned}
\log &p(y \mid \pi_0, \alpha, \beta)
= \mathbbm{1}_{\{y=0\}}\, \log \pi_0 \\[4pt]
&\quad +\, \mathbbm{1}_{\{y>0\}}\,
\big[\log(1 - \pi_0) + \log \mathrm{Gamma}(y; \alpha, \beta)\big],
\end{aligned}
\label{eq:zig_loglikelihood}
\end{equation*}
where $\mathbbm{1}$ is the indicator function.
%
%

Let $p = \mathbbm{1}_{\{1 - \pi_0 \ge 0.5\}}$ indicate nonzero rainfall, and the Gamma component have mean and variance
$\mu_\Gamma = \frac{\alpha}{\beta}, 
\qquad
\sigma_\Gamma^2 = \frac{\alpha}{\beta^2}.
$  
Then, mean and variance of the ZIG distribution are (derivation in Appendix~\ref{app:zig}):
\begin{equation*}
    \mathbb{E}[Y] = p\,\mu_\Gamma, \quad
    \mathrm{Var}[Y] = p\,\sigma_\Gamma^2 + p(1 - p)\mu_\Gamma^2.
\end{equation*}

%% file: sections/4_experiments.tex
\section{Experiments}
\label{sec:experiments}
DropsToGrid is implemented in PyTorch \citep{pytorch} and PyTorch Lightning \citep{pytorch_lightning} (details in Appendix~\ref{app:training}). The best checkpoint by validation loss is used for evaluation. The 192K-parameter model is trained on an NVIDIA H100 GPU for up to 3 hours. Results are averaged over three seeds, reporting mean and standard deviation.

We measure Critical Success Index (CSI)~\cite{csi}: evaluates prediction accuracy across multiple intensity thresholds, emphasizing performance in fields dominated by no-rain events \cite{metnet3, earthformer}; Fraction Skill Score (FSS)~\cite{fss}: measures spatial and intensity agreement, mitigating small displacement and shape errors; Frequency Bias Index (FBI)~\cite{fbi}: quantifies systematic over- or underprediction of rain occurrences (details and extended results in Appendix~\ref{app:metrics}).

We evaluate densification of sparse 1-hour rainfall accumulations (in mm) derived from crowd-sourced PWS data. Measurements are projected onto a 4 km grid covering Denmark, forming a $96 \times 96$ patch.
The dataset spans January--December 2024 and is divided into 12-, 2-, and 2-day periods for training, validation, and test, with 12-hour blackout preventing temporal leakage. During training, each patch is randomly masked to retain 30--50\% of PWS pixels, simulating sparsity and increasing effective sample size.

To assess spatial generalization, we evaluate on a holdout set of 20\% randomly selected pixels with stations. These pixels are excluded from all model inputs during both training and validation, ensuring the model is tested on locations it has not seen before. For independent benchmarking, we also draw upon official Danish Meteorological Institute (DMI) rainfall observations from SYNOP stations with research-grade gauges and tipping-bucket sensors from local wastewater utilities \cite{pluvio}.

We compare against several operational or reanalysis rainfall products, treated as baselines rather than ground truth: OPERA~\cite{opera_1, opera_2} radar-derived accumulations, RainViewer~\footnote{\url{https://www.rainviewer.com}} radar reflectivity, IMERG~\cite{imerg} satellite retrievals, ERA5~\cite{era5, era5_2} reanalysis, and DMI's gridded \textit{Climate}~\cite{climategrid} dataset. Further details in Appendix~\ref{app:baselines}.
%
%
\begin{table*}[!t]
  \centering
  \caption{Performance comparison against operational estimators on holdout set of PWS stations and research-quality SYNOP stations. Our proposed method achieves superior performance across all metrics. Best results are shown in \textbf{bold}, and second-best in \textit{italics}.}
  \label{tab:baselines_stations}
  \begin{tabular}{lcccccc}
    \toprule
    \multirow{2}{*}{Method} & \multicolumn{3}{c}{PWS holdout} & \multicolumn{3}{c}{SYNOP Stations} \\
    \cmidrule(lr){2-4} \cmidrule(lr){5-7}
    & CSI $\uparrow$ & FSS $\uparrow$ & FBI $\approx1$ & CSI $\uparrow$ & FSS $\uparrow$ & FBI $\approx1$ \\
    \midrule
    OPERA & 0.298 & 0.526 & 0.596 & \textit{0.323} & \textit{0.551} & \textit{0.688} \\
    RainViewer & 0.292 & 0.546 & 1.491 & 0.304 & 0.525 & 2.231 \\
    IMERG & 0.202 & 0.449 & \textit{1.123} & 0.194 & 0.425 & 1.315 \\
    ERA5 & 0.204 & 0.390 & 0.638 & 0.203 & 0.363 & 0.632 \\
    Climate & \textit{0.437} & \textit{0.740} & 0.836 & - & - & - \\
    DropsToGrid & \textbf{0.532 ± 0.002} & \textbf{0.819 ± 0.000} & \textbf{0.877 ± 0.019} & \textbf{0.551 ± 0.006} & \textbf{0.795 ± 0.007} & \textbf{0.995 ± 0.030} \\
    \bottomrule
  \end{tabular}
\end{table*}

\begin{table*}[!t]
  \centering
  \caption{Pairwise comparison of gridded products using CSI. Although no product represents a true ground truth, the higher similarity of DropsToGrid to other products suggests greater robustness and consistency. Best result is shown in \textbf{bold}, and second-best in \textit{italics}.}
  \label{tab:dense_operational}
  \begin{tabular}{lcccccc}
    \toprule
    \quad CSI $\uparrow$ & OPERA & RainViewer & IMERG & ERA5 & Climate & DropsToGrid \\
    \midrule
    OPERA & - & \textit{0.360} & 0.207 & 0.153 & 0.294 & 0.327 ± 0.001 \\
    RainViewer & \textit{0.360} & - & 0.230 & 0.163 & 0.286 & 0.332 ± 0.001 \\
    IMERG & 0.207 & 0.230 & - & 0.153 & 0.208 & 0.217 ± 0.002 \\
    ERA5 & 0.153 & 0.163 & 0.153 & - & 0.219 & 0.205 ± 0.001 \\
    Climate & 0.294 & 0.286 & 0.208 & 0.219 & - & \textbf{0.489 ± 0.003} \\
    DropsToGrid & 0.327 ± 0.001 & 0.332 ± 0.001 & 0.217 ± 0.002 & 0.205 ± 0.001 & \textbf{0.489 ± 0.003} & - \\
    \bottomrule
  \end{tabular}
\end{table*}

Table~\ref{tab:baselines_stations} compares gridded products against the PWS holdout set and DMI SYNOP stations. Climate is excluded from the SYNOP evaluation to avoid leakage, as these observations are used to generate its grid. DropsToGrid outperforms all operational models, achieving the highest CSI (best detection), FSS (most accurate spatial structure), and FBI closest to 1 (least bias). OPERA underestimates rainfall, RainViewer overestimates, and IMERG and ERA5 show limited performance. Table~\ref{tab:dense_operational} presents pairwise CSI scores among the different estimators. Beyond providing uncertainty estimates, DropsToGrid demonstrates robust performance and superior extrapolation to unseen areas than Climate, as reflected by stronger performance on PWS holdout and higher agreement with other products. Figure~\ref{fig:comparison} illustrates these comparisons (more cases in Appendix~\ref{app:visualizations}).

\begin{figure}
  \centering
   \includegraphics[width=\linewidth]{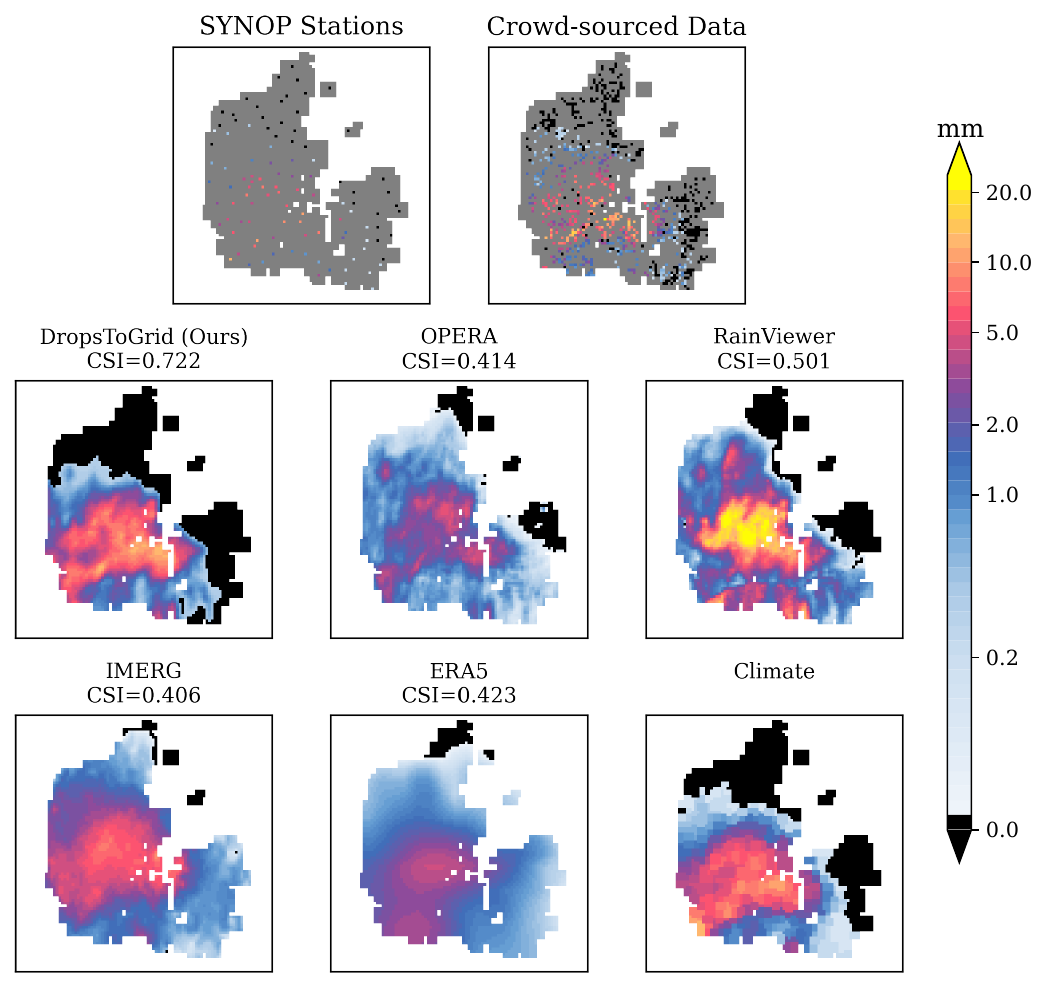}
   \caption{Hourly rainfall accummulation estimates vs test stations (No CSI on Climate as SYNOP stations input to gridded product).}
   \label{fig:comparison}
\end{figure}

\textbf{Comparison of Deep Learning models.} 
We compare DropsToGrid with leading deep learning densification models, including ConvCNP and SwinTNP variants from Gridded TNPs ~\cite{gridded-tnp}. Variants use either station temporal sequences without radar (MM) or single-timestep stations with radar (OOTG). ConvCNP employs SetConv and U-Net-style convolutions (2D or 3D), merging radar pixel-wise if available, without specialized fusion. SwinTNP projects high-resolution inputs into a grid via attention, applies a Swin Transformer~\cite{swin} in the dense latent space, and decodes back to high resolution. All models are retrained with the original code and loss from Section~\ref{ssec:loss}. Since these NP-based models provide uncertainty estimates, we also evaluate the Continuous Ranked Probability Score (CRPS) \cite{crps} to assess sharpness and calibration. Table~\ref{tab:dl_baselines} shows that DropsToGrid outperforms all baselines across metrics and datasets, effectively fusing data sources and timesteps, crucial for densifying sparse PWS observations. Using radar (OOTG) vs station history (MM) is not uniformly beneficial: ConvCNP improves detection (CSI) and calibration (CRPS) but slightly loses spatial consistency (FSS), while SwinTNP consistently benefits from multimodal input. ConvCNP generally outperforms SwinTNP due to translation-equivariance, yet DropsToGrid achieves the best balance of spatial performance and uncertainty calibration. Further details in Appendix~\ref{app:dl_baselines}.
\begin{table*}
  \centering
  \caption{Performance comparison of deep learning baselines on holdout set of crowd-sourced stations and quality-controlled SYNOP stations. Our proposed method achieves superior performance across all metrics. Best results are shown in \textbf{bold}, and second-best in \textit{italics}.}
  \label{tab:dl_baselines}
  \begin{tabular}{lcccccc}
    \toprule
    \multirow{2}{*}{Method} & \multicolumn{3}{c}{PWS holdout} & \multicolumn{3}{c}{SYNOP Stations} \\
    \cmidrule(lr){2-4} \cmidrule(lr){5-7}
    & CSI $\uparrow$ & FSS $\uparrow$ & CRPS $\downarrow$ & CSI $\uparrow$ & FSS $\uparrow$ & CRPS $\downarrow$ \\
    \midrule
    MM\_ConvCNP & 0.484 ± 0.012 & \textit{0.769 ± 0.035} & 0.029 ± 0.000 & 0.507 ± 0.012 & \textit{0.760 ± 0.020} & \textit{0.025 ± 0.000} \\
    MM\_SwinTNP & 0.407 ± 0.002 & 0.676 ± 0.003 & 0.034 ± 0.000 & 0.424 ± 0.005 & 0.666 ± 0.008 & 0.030 ± 0.000 \\
    OOTG\_ConvCNP & \textit{0.486 ± 0.007} & 0.748 ± 0.013 & \textit{0.027 ± 0.000} & \textit{0.522 ± 0.015} & 0.754 ± 0.021 & \textbf{0.023 ± 0.000} \\
    OOTG\_SwinTNP & 0.433 ± 0.001 & 0.711 ± 0.006 & 0.032 ± 0.000 & 0.457 ± 0.005 & 0.700 ± 0.007 & 0.028 ± 0.000 \\
    DropsToGrid & \textbf{0.532 ± 0.002} & \textbf{0.819 ± 0.000} & \textbf{0.026 ± 0.000} & \textbf{0.551 ± 0.006} & \textbf{0.795 ± 0.007} & \textbf{0.023 ± 0.000} \\
    \bottomrule
  \end{tabular}
\end{table*}

\textbf{Ablation.} 
DropsToGrid encodes each timestep and source independently, merging their high-dimensional latent representations at the U-Net's bottleneck with temporal attention to combine timesteps, cross-modal attention to integrate radar, and a translation-equivariant fusion transformer. We compare this design against a standard convolutional bottleneck that jointly processes all sources and timesteps. Also, all models are trained without any input stations in the target to prevent direct input-output mapping and handle noisy observations. Results show that both specialized fusion bottleneck and exclusion of input stations in the targets are critical, with 4\% and 8\% drop in CSI against SYNOP stations otherwise. Further details in Appendix~\ref{app:ablation}.

\begin{figure}
  \centering
  \includegraphics[width=0.35\linewidth]{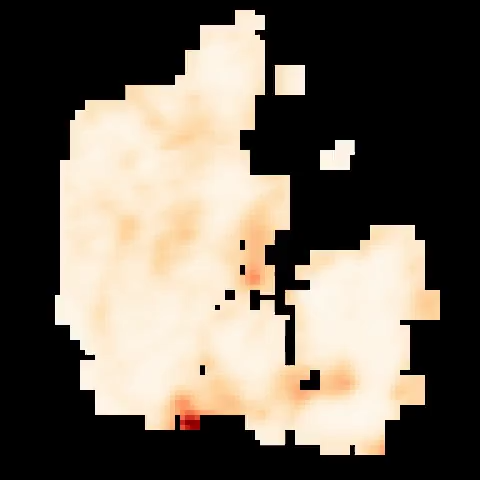}
  \includegraphics[width=0.35\linewidth]{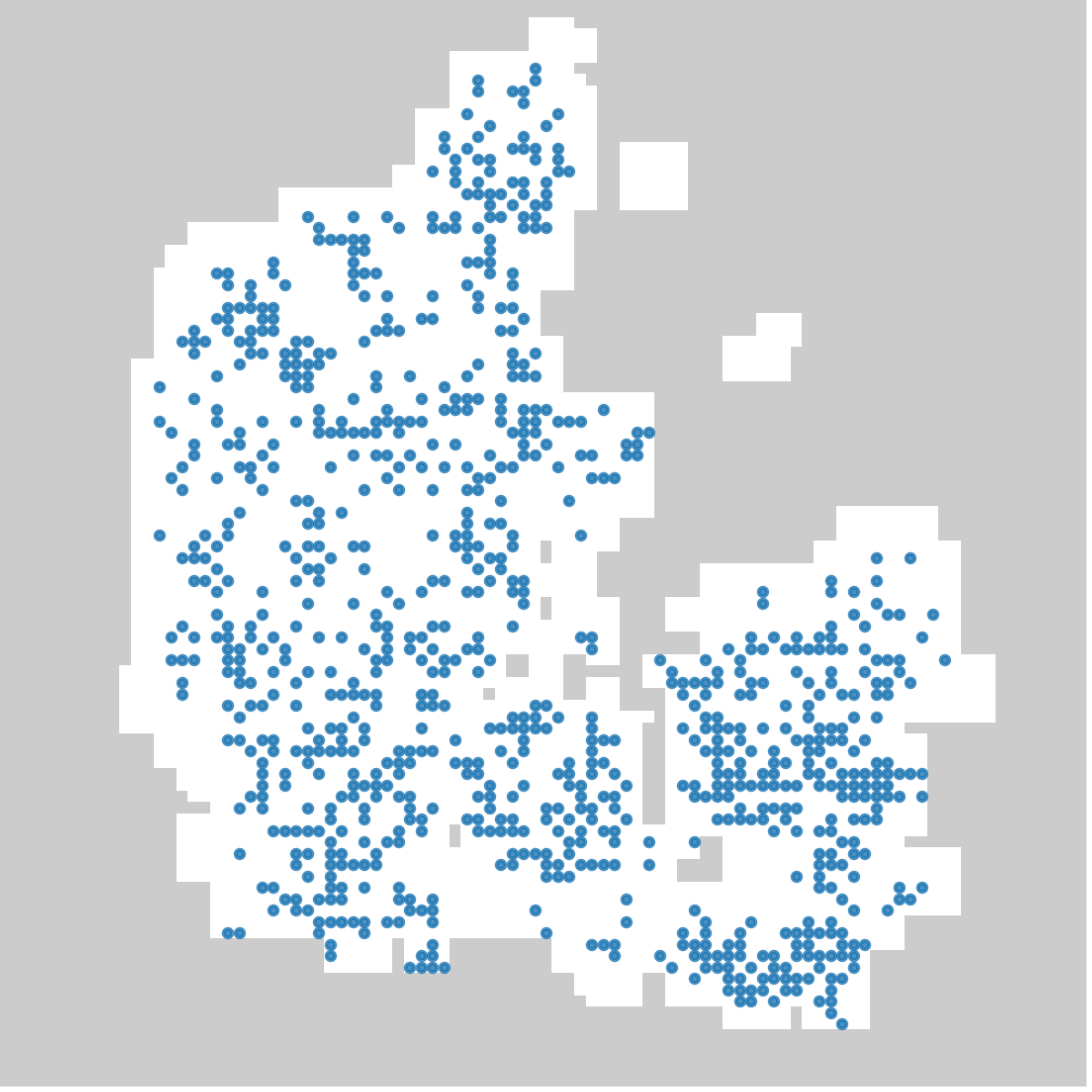}
  \caption{Overall model uncertainty (left) vs PWS stations (right).}
  \label{fig:uncertainty}
\end{figure}

\textbf{Uncertainty maps.} 
Uncertainty maps are valuable for applications such as optimizing station placement~\cite{rainfall_kriging} or detecting inconsistent sensors. Fig.~\ref{fig:uncertainty} shows DropsToGrid's uncertainty heatmap and the distribution of PWS stations. Uncertainty is estimated per prediction and is typically higher in regions with rainfall, since non-rain areas are easier to reconstruct. Averaging across the test set reveals consistent spatial patterns: uncertainty is non-negligible at PWS locations, reflecting DropsToGrid's awareness of observation noise, and it increases in regions farther from stations, highlighting areas where predictions are less certain.

\textbf{Station analysis.} 
Although densifying rainfall observations from PWS stations is challenging, even in high-density regions, due to observation noise and highly localized rainfall patterns, DropsToGrid performs strongly under these conditions. Moreover, using only 5\% of the stations, it still outperforms operational estimators by over 24\% in CSI against SYNOP stations, demonstrating its strong potential and effectiveness even in low-density regions. The best results are obtained when all available stations are used, highlighting the importance of leveraging the full observational network. Figures and details are in Appendix~\ref{app:stations}.

\textbf{Europe-wide cross-regional evaluation.} 
We evaluate DropsToGrid across Europe for the full year of 2025, entirely unseen during training, spanning diverse climatic and topographic conditions. A model trained solely in Denmark demonstrates strong cross-region generalization, while an EU-wide model further improves performance on the unseen year 2025, remaining compact and scalable. DropsToGrid effectively ignores noise from PWS observations, leverages radar in sparse regions, and corrects systematic biases, enhancing reliability. Despite lower data quality on the global PWS network and sparser stations compared to Danish training data, DropsToGrid consistently outperforms operational baselines and other learned models. Further details are provided in Appendix~\ref{app:europe}.

%% file: sections/5_conclusion.tex
\section{Conclusion}
We present DropsToGrid, a Neural Process–based approach for probabilistic rainfall densification that effectively fuses sparse, noisy station measurements with dense radar context to produce high-resolution rainfall fields with calibrated uncertainty. Our experiments on real-world datasets demonstrate that DropsToGrid captures both local and large-scale rainfall patterns, outperforming operational and deep learning baselines, and remains robust even in low-density observation scenarios and regions and time periods outside the training domain. Beyond improving rainfall estimation, DropsToGrid’s uncertainty-aware framework provides actionable insights for sensor placement and data quality assessment. Looking forward, this approach can be extended by integrating additional meteorological variables not only as inputs but also as predictive targets, moving toward a general probabilistic data assimilation framework.

%% file: sections/appendix/01_zig.tex
\section{Derivation of ZIG mean and variance}
\label{app:zig}

The model outputs a zero-rain probability $\pi_0$, from which we define a deterministic per-sample rain indicator
\[
p \;=\; \mathbbm{1}_{\{1-\pi_0 \ge 0.5\}},
\]
so that $p=1$ denotes predicted nonzero rainfall and $p=0$ otherwise.  
Given this indicator, the Zero-Inflated Gamma (ZIG) variable $Y$ is
\[
Y \mid (p=0) = 0, \qquad 
Y \mid (p=1) \sim \textrm{Gamma}(\alpha,\beta),
\]
with Gamma mean and variance
\[
\mu_\Gamma = \frac{\alpha}{\beta}, 
\qquad
\sigma_\Gamma^2 = \frac{\alpha}{\beta^2}.
\]

Because $p \in \{0,1\}$ is deterministic for each sample, conditioning removes all mixture uncertainty: $Y$ is either identically zero or drawn from a Gamma distribution.  
Applying the law of total expectation and the law of total variance, respectively:
\begin{align*}
\mathbb{E}[Y] &= \mathbb{E}\big[\mathbb{E}[Y \mid p]\big]
               = (1-p)\cdot 0 + p \cdot \mu_\Gamma
               = p\,\mu_\Gamma, \\
\mathrm{Var}[Y] &= \mathbb{E}\!\left[\mathrm{Var}(Y \mid p)\right]
                 + \mathrm{Var}\!\left(\mathbb{E}[Y \mid p]\right).
\end{align*}

The second term vanishes because $\mathbb{E}[Y \mid p]$ is constant for the fixed value of $p$ in a given sample. Therefore,
\[
\mathrm{Var}[Y] 
= p\,\sigma_\Gamma^2.
\]

\paragraph{Remark.}  
The more general ZIG variance expression
\[
\mathrm{Var}[Y] = p\,\sigma_\Gamma^2 + p(1-p)\mu_\Gamma^2
\]
applies when $p \in [0,1]$ represents a \emph{probabilistic mixture weight}, but the deterministic case with indicator $p$ used here yields the simplified variance $p\,\sigma_\Gamma^2$.

\newpage

%% file: sections/appendix/02_training.tex
\section{Training and dataset details}
\label{app:training}

DropsToGrid employs a U-Net of depth 3 with a kernel size of 3, 32 channels, a bottleneck dropout of 0.1, and transformer blocks of depth 2 with 8 heads of dimension 8. The model contains a total of 192K parameters. 

It is trained for up to 50K steps with a batch size of 32, and validation is performed every 1,000 steps. Training uses the AdamW optimizer with a cosine-annealed learning rate starting at \(3\times10^{-4}\), weight decay \(1\times10^{-4}\), betas \((0.9, 0.999)\), and an EMA decay of 0.999. On an NVIDIA H100 PCIe GPU, training requires approximately 3 hours.

The dataset spans January-December 2024 and is divided into training, validation, and test periods of 12, 2, and 2 days, respectively, separated by a 12-hour blackout to prevent temporal leakage. This results in 34,287 training, 911 validation, and 914 test patches. 

The target region of $384 \times 384~\mathrm{km}^2$ is projected onto a 4km/px grid. When multiple stations fall within the same grid cell, the median value is used to mitigate the influence of outliers. Across all samples, up to 902 possible pixels are active in the training set, varying from 797 on January 1st to 884 on December 31st due to station additions, relocations, or temporary outages in the PWS network. The region of interest includes 4,003 land and near-coastal pixels for rainfall estimation. A total of 104 grid cells contain SYNOP stations, with most stations concentrated in urban areas, leading to uneven spatial coverage.

\newpage

%% file: sections/appendix/03_metrics.tex
\section{Metrics and extended results}
\label{app:metrics}

We evaluate predictions using Critical Success Index (CSI), Fraction Skill Score (FSS), Frequency Bias Index (FBI), and Continuous Ranked Probability Score (CRPS). Additionally, we report Mean Absolute Error (MAE) and Mean Squared Error (MSE), noting that these are sensitive to the prevalence of no-rain events and may be less informative for highly skewed precipitation distributions \cite{metnet3}.

\paragraph{Critical Success Index (CSI)}
CSI, or Threat Score, measures event detection accuracy by comparing correctly predicted precipitation events to all predicted or observed events. It balances misses and false alarms, making it especially useful for rare-event forecasting such as heavy rainfall. CSI ranges from 0 (no skill) to 1 (perfect prediction).
\begin{equation}
\text{CSI} = \frac{\text{TP}}{\text{TP}+\text{FP}+\text{FN}},
\end{equation}
where TP, FP, and FN are true positives, false positives, and false negatives, respectively, using thresholds of 0.2, 1, 2, 5, and 10 mm/h.

\paragraph{Fraction Skill Score (FSS)}
FSS evaluates the spatial accuracy of predictions by comparing the fraction of predicted and observed positive events within a local neighborhood, rather than pointwise. It accounts for small spatial displacements, making it suitable for high-resolution precipitation estimation. FSS ranges from 0 (no spatial agreement) to 1 (perfect prediction).

\begin{equation}
\text{FSS} = 1 - \frac{\sum_{i=1}^{H}\sum_{j=1}^{W}(F_{i,j}-O_{i,j})^2}{\sum_{i=1}^{H}\sum_{j=1}^{W}F_{i,j}^2 + \sum_{i=1}^{H}\sum_{j=1}^{W}O_{i,j}^2},
\end{equation}
where \(F_{i,j}\) and \(O_{i,j}\) are the fractions of predicted and observed positives in the neighborhood of pixel \((i,j)\). FSS is computed at thresholds 0.2, 1, 2, 5, and 10 mm/h, and for neighborhood sizes of 2, 10, and 20 pixels.

\paragraph{Frequency Bias Index (FBI)}
FBI quantifies systematic over- or underprediction of events. Values greater than 1 indicate overprediction, while values below 1 indicate underprediction. A bias of 1 indicates that the predicted frequency matches the observed frequency, though not necessarily the spatial or intensity accuracy.

\begin{equation}
\text{FBI} = \frac{\text{TP}+\text{FP}}{\text{TP}+\text{FN}},
\end{equation}
evaluated at the same precipitation thresholds as CSI and FSS (0.2, 1, 2, 5, and 10 mm/h).

\paragraph{Continuous Ranked Probability Score (CRPS)}
CRPS evaluates the accuracy of a probabilistic forecast by comparing the predicted cumulative distribution function (CDF) to the observed outcome. Lower values indicate better forecast sharpness, calibration, and reliability, rewarding predictions that assign high probability to the correct rainfall intensity. Computation follows closed-form solutions for Gamma distributions~\cite{crps_gamma}. We report CRPS for the probabilistic models (i.e., ablations and deep learning baselines).

For a Zero-Inflated Gamma (ZIG) forecast, which models rainfall $R$ as a mixture of a point mass at zero and a continuous Gamma for positive rainfall, CRPS is computed as:

\begin{equation}
\text{CRPS}_{\text{ZIG}} = 
\begin{cases} 
p_{\text{nonzero}} \cdot \text{CRPS}_{\Gamma}(y=0), & R \le 0.2 \\
p_{\text{zero}} \cdot R + p_{\text{nonzero}} \cdot \text{CRPS}_{\Gamma}(R), & R > 0.2
\end{cases}
\end{equation}

where $p_{\text{nonzero}} \in \{0,1\}$ is the binarized predicted probability of nonzero rainfall, $p_{\text{zero}} = 1 - p_{\text{nonzero}}$, and $\text{CRPS}_{\Gamma}$ is the closed-form CRPS for the Gamma component~\cite{crps_gamma}:

\begin{align}
\text{CRPS}_{\Gamma}(y) &= y \left(2 F_\Gamma(y) - 1\right) \nonumber - \frac{\alpha}{\beta} \left(2 F_{\Gamma,\alpha+1}(y) - 1\right) \nonumber \\
&\quad - \frac{\alpha}{\beta \pi} B\Big(\alpha+\frac{1}{2}, \frac{1}{2}\Big)
\end{align}

with $F_\Gamma$ the CDF of the Gamma distribution with parameters $(\alpha, \beta)$, $F_{\Gamma,\alpha+1}$ the CDF with shape $\alpha+1$, and $B(\cdot, \cdot)$ the Beta function.

\paragraph{Extended results}
Tables~\ref{tab_app:metrics_1}–\ref{tab_app:metrics_6} present a more detailed evaluation of DropsToGrid and the gridded baselines against SYNOP stations, including overall metrics, threshold-specific results for CSI, FBI, and FSS, as well as different neighborhood sizes for FSS.

\begin{table*}[h!]
\centering
\caption{Performance comparison across metrics against operational estimators on research-quality SYNOP stations. DropsToGrid achieves superior performance across all metrics. Best results are shown in \textbf{bold}, and second-best in \textit{italics}.}
\label{tab_app:metrics_1}
\begin{tabular}{llllll}
\toprule
& CSI $\uparrow$ & FSS $\uparrow$ & FBI $\approx1$ & MAE $\downarrow$ & MSE $\downarrow$ \\
\midrule
OPERA & \textit{0.323} & \textit{0.551} & \textit{0.688} & \textit{0.062} & \textit{0.118} \\
RainViewer & 0.304 & 0.525 & 2.231 & 0.089 & 0.599 \\
IMERG & 0.194 & 0.425 & 1.315 & 0.111 & 0.261 \\
ERA5 & 0.203 & 0.363 & 0.632 & 0.084 & 0.174 \\
DropsToGrid & \textbf{0.551 ± 0.006} & \textbf{0.795 ± 0.007} & \textbf{0.995 ± 0.030} & \textbf{0.037 ± 0.001} & \textbf{0.070 ± 0.005} \\
\bottomrule
\end{tabular}
\end{table*}

\begin{table*}[h!]
\centering
\caption{Performance comparison across thresholds for CSI against operational estimators on research-quality SYNOP stations. \mbox{DropsToGrid} achieves superior performance across all metrics. Best results are shown in \textbf{bold}, and second-best in \textit{italics}.}
\label{tab_app:metrics_2}
\begin{tabular}{llllll}
\toprule
& \multicolumn{5}{c}{CSI $\uparrow$} \\
\cmidrule(lr){2-6}
& 0.2 mm/h & 1.0 mm/h & 2.0 mm/h & 5.0 mm/h & 10.0 mm/h \\
\midrule
OPERA & \textit{0.501} & \textit{0.460} & \textit{0.375} & \textit{0.156} & \textit{0.125} \\
RadarViewer & 0.498 & 0.446 & 0.367 & 0.180 & 0.029 \\
IMERG & 0.280 & 0.284 & 0.259 & 0.102 & 0.044 \\
ERA5 & 0.369 & 0.367 & 0.281 & 0.000 & 0.000 \\
DropsToGrid & \textbf{0.674 ± 0.002} & \textbf{0.656 ± 0.002} & \textbf{0.639 ± 0.007} & \textbf{0.444 ± 0.003} & \textbf{0.342 ± 0.024} \\
\bottomrule
\end{tabular}
\end{table*}

\begin{table*}[h!]
\centering
\caption{Performance comparison across thresholds for FBI against operational estimators on research-quality SYNOP stations. \mbox{DropsToGrid} achieves superior performance across all metrics. Best results are shown in \textbf{bold}, and second-best in \textit{italics}.}
\label{tab_app:metrics_3}
\begin{tabular}{llllll}
\toprule
& \multicolumn{5}{c}{FBI $\approx1$} \\
\cmidrule(lr){2-6}
& 0.2 mm/h & 1.0 mm/h & 2.0 mm/h & 5.0 mm/h & 10.0 mm/h \\
\midrule
OPERA & \textit{1.093} & 0.794 & 0.657 & \textit{0.512} & 0.385 \\
RadarViewer & \textbf{0.973} & \textit{1.059} & \textit{1.076} & 2.122 & 5.923 \\
IMERG & 1.429 & 1.459 & 1.364 & 1.517 & \textbf{0.808} \\
ERA5 & 1.346 & 1.081 & 0.731 & 0.000 & 0.000 \\
DropsToGrid & 0.887 ± 0.005 & \textbf{1.054 ± 0.005} & \textbf{1.041 ± 0.020} & \textbf{1.238 ± 0.038} & \textit{0.756 ± 0.101} \\
\bottomrule
\end{tabular}
\end{table*}

\begin{table*}[h!]
\centering
\caption{Performance comparison across thresholds for FSS (neighborhood size: 2px) against operational estimators on research-quality SYNOP stations. DropsToGrid achieves superior performance across all metrics. Best results are shown in \textbf{bold}, and second-best in \textit{italics}.}
\label{tab_app:metrics_4}
\begin{tabular}{llllll}
\toprule
& \multicolumn{5}{c}{FSS (size: 2px) $\uparrow$} \\
\cmidrule(lr){2-6}
& 0.2 mm/h & 1.0 mm/h & 2.0 mm/h & 5.0 mm/h & 10.0 mm/h \\
\midrule
OPERA & \textit{0.671} & \textit{0.635} & \textit{0.550} & 0.266 & \textit{0.220} \\
RadarViewer & 0.669 & 0.623 & 0.543 & \textit{0.310} & 0.056 \\
IMERG & 0.440 & 0.444 & 0.415 & 0.184 & 0.084 \\
ERA5 & 0.542 & 0.539 & 0.443 & 0.000 & 0.000 \\
DropsToGrid & \textbf{0.809 ± 0.001} & \textbf{0.796 ± 0.002} & \textbf{0.784 ± 0.005} & \textbf{0.622 ± 0.003} & \textbf{0.510 ± 0.031} \\
\bottomrule
\end{tabular}
\end{table*}

\begin{table*}[h!]
\centering
\caption{Performance comparison across thresholds for FSS (neighborhood size: 10px) against operational estimators on research-quality SYNOP stations. DropsToGrid achieves superior performance across all metrics. Best results are shown in \textbf{bold}, and second-best in \textit{italics}.}
\label{tab_app:metrics_5}
\begin{tabular}{llllll}
\toprule
& \multicolumn{5}{c}{FSS (size: 10px) $\uparrow$} \\
\cmidrule(lr){2-6}
& 0.2 mm/h & 1.0 mm/h & 2.0 mm/h & 5.0 mm/h & 10.0 mm/h \\
\midrule
OPERA & \textit{0.780} & \textit{0.756} & 0.679 & 0.338 & \textit{0.261} \\
RadarViewer & 0.777 & 0.752 & \textit{0.680} & \textit{0.419} & 0.055 \\
IMERG & 0.549 & 0.564 & 0.535 & 0.284 & 0.214 \\
ERA5 & 0.628 & 0.640 & 0.557 & 0.000 & 0.000 \\
DropsToGrid & \textbf{0.900 ± 0.001} & \textbf{0.897 ± 0.001} & \textbf{0.892 ± 0.004} & \textbf{0.782 ± 0.001} & \textbf{0.597 ± 0.039} \\
\bottomrule
\end{tabular}
\end{table*}

\begin{table*}[h!]
\centering
\caption{Performance comparison across thresholds for FSS (neighborhood size: 20px) against operational estimators on research-quality SYNOP stations. DropsToGrid achieves superior performance across all metrics. Best results are shown in \textbf{bold}, and second-best in \textit{italics}.}
\label{tab_app:metrics_6}
\begin{tabular}{llllll}
\toprule
& \multicolumn{5}{c}{FSS (size: 20px) $\uparrow$} \\
\cmidrule(lr){2-6}
& 0.2 mm/h & 1.0 mm/h & 2.0 mm/h & 5.0 mm/h & 10.0 mm/h \\
\midrule
OPERA & \textit{0.843} & 0.816 & 0.740 & 0.418 & 0.289 \\
RadarViewer & 0.839 & \textit{0.829} & \textit{0.757} & \textit{0.488} & 0.076 \\
IMERG & 0.628 & 0.662 & 0.636 & 0.385 & \textit{0.359} \\
ERA5 & 0.700 & 0.730 & 0.664 & 0.000 & 0.000 \\
DropsToGrid & \textbf{0.943 ± 0.000} & \textbf{0.946 ± 0.001} & \textbf{0.943 ± 0.003} & \textbf{0.860 ± 0.003} & \textbf{0.649 ± 0.040} \\
\bottomrule
\end{tabular}
\end{table*}

\clearpage

%% file: sections/appendix/04_baselines.tex
\clearpage
\section{Baseline gridded products}
\label{app:baselines}

For evaluation, we use operational and reanalysis gridded rainfall products as reference baselines rather than ground truth. To ensure comparability, all baselines are resampled to a uniform 4 km grid using bilinear interpolation and converted to hourly rainfall accumulations (mm). Further details on each gridded baseline product are provided below.

\paragraph{OPERA Odyssey rainfall accumulation.} The EUMETNET OPERA program \cite{opera_1, opera_2} provides pan-European radar composites combining data from over 160 national radars. The 1-hour accumulation product has 2 km resolution and 15-minute updates. Rain rate is derived from reflectivity using the Marshall-Palmer $Z=aR^b$ relation ($a=200$, $b=1.6$) \cite{marshall_palmer}. Quality filtering employs anomaly removal \cite{bropo}, clutter \cite{hac}, and beam-blockage \cite{beamb} corrections. Maximum accumulation is set to 100 mm.

\paragraph{RainViewer reflectivity.} RainViewer~\footnote{\url{https://www.rainviewer.com}} provides near-real time radar composites every 10 minutes at 2km resolution, aggregating over 1,000 stations worldwide. Reflectivity (dBZ) is clipped to [-1, 64] and converted to intensity (mm/h) via Marshall-Palmer \cite{marshall_palmer}. Hourly accumulations are computed by averaging the six 10-min intensity maps.

\paragraph{IMERG.} The GPM IMERG product \cite{imerg} merges passive microwave, infrared, and radar data from multiple satellites to provide global rainfall estimates at 0.1° and 30-min intervals. We use the Final Run V07 dataset, bias-corrected to match rain-gauge climatologies. Hourly accumulations are obtained by averaging the two 30-min frames per hour.

\paragraph{ERA5.} ERA5 \cite{era5, era5_2} is ECMWF's global reanalysis, combining a frozen numerical model with a consistent data assimilation scheme. Although it does not directly assimilate European rainfall observations, estimates are derived from physically consistent atmospheric fields (humidity, geopotential, winds). ERA5 provides total rainfall, including convective, large-scale, and evaporative effects, at 0.25° resolution and hourly frequency.

\paragraph{Climate (DMI gridded product).} The DMI \textit{Climate} dataset \cite{climategrid} provides hourly gridded rainfall at 10 km resolution. Fields are generated using Inverse Distance Weighting (IDW) interpolation of quality-controlled gauges, followed by Gaussian smoothing and a coastal-inland correction to adjust distance metrics across land-sea boundaries. The gauge network used to generate the gridded product includes the SYNOP stations previously mentioned.

%% file: sections/appendix/05_visualizations.tex
\section{Visualizations}
\label{app:visualizations}

\begin{figure*}[p]
  \centering
   \includegraphics[width=0.9\linewidth]{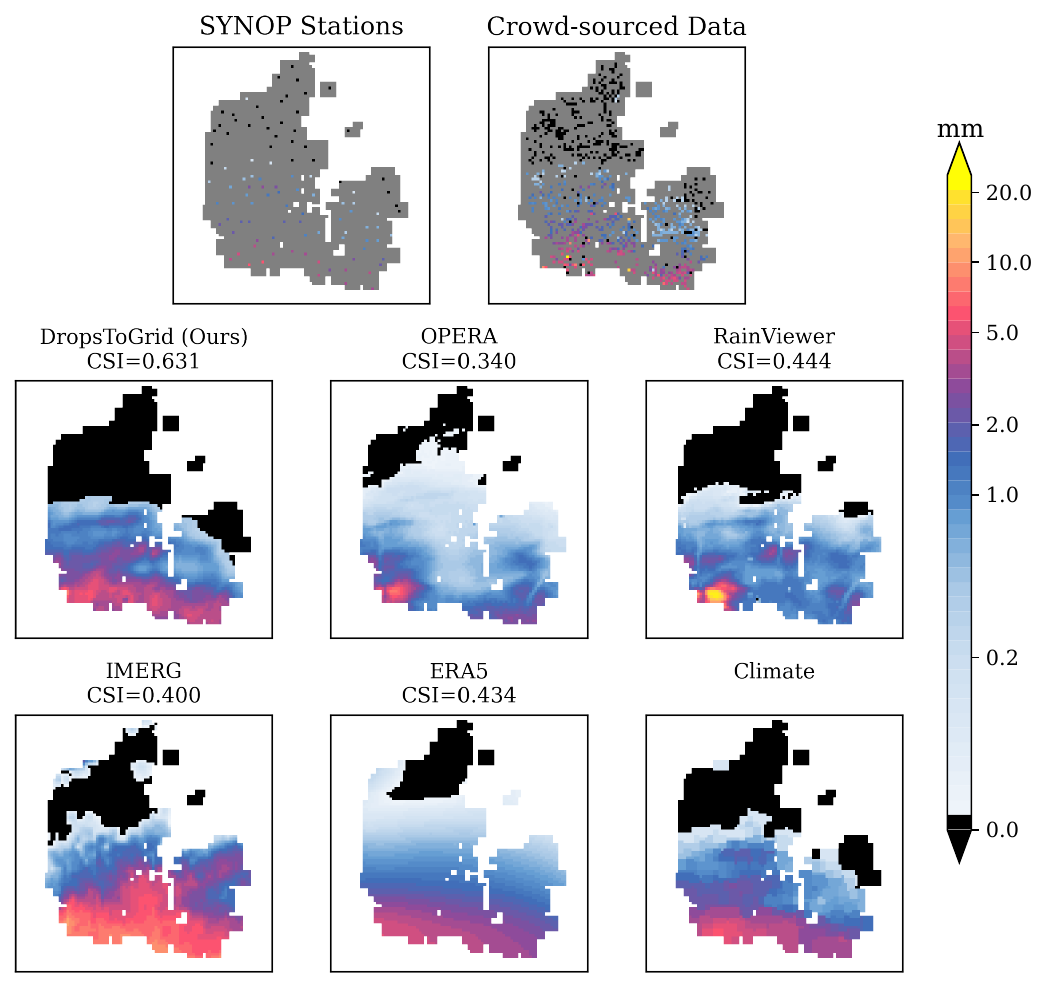}
   \caption{Comparison of rainfall estimators against research-quality SYNOP stations.}
   \label{fig_app:viz_1}
\end{figure*}

\begin{figure*}[p]
  \centering
   \includegraphics[width=0.9\linewidth]{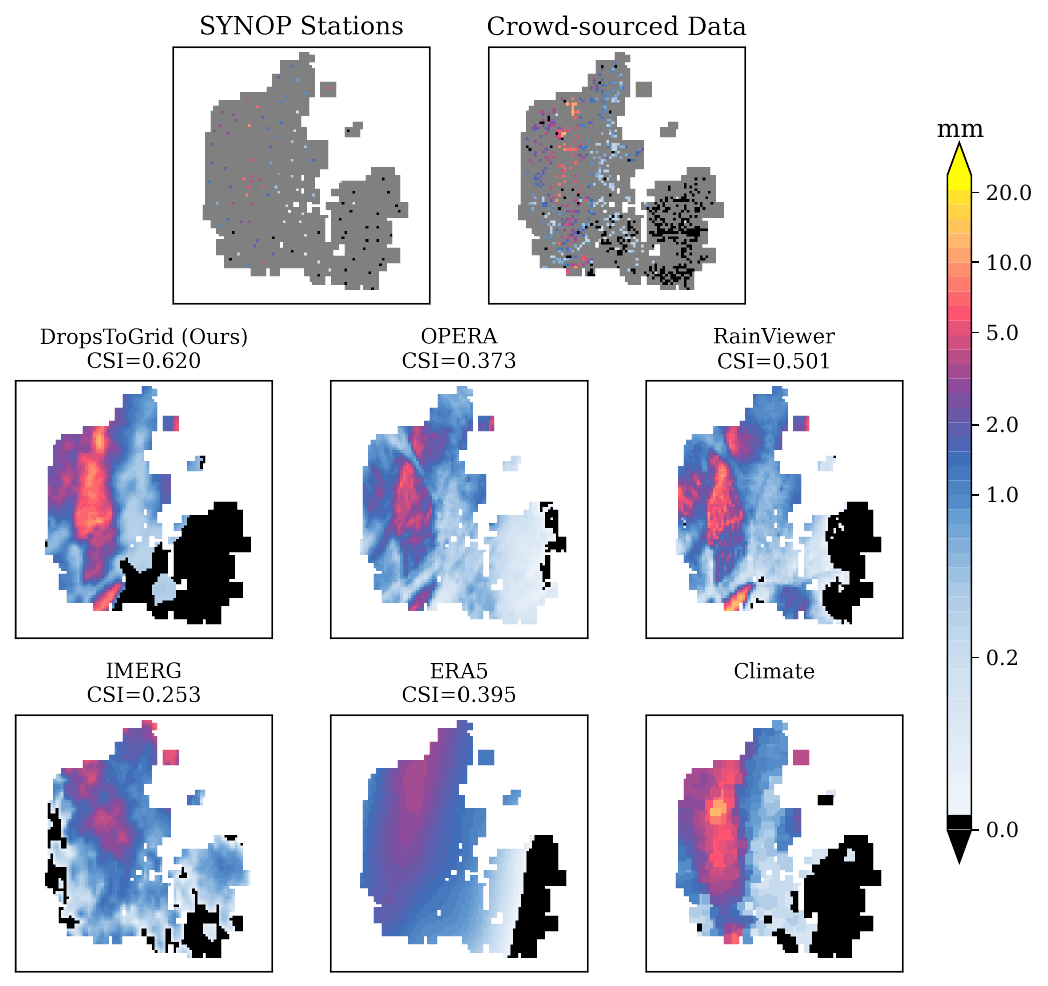}
   \caption{Comparison of rainfall estimators against research-quality SYNOP stations.}
   \label{fig_app:viz_2}
\end{figure*}

\begin{figure*}[p]
  \centering
   \includegraphics[width=0.9\linewidth]{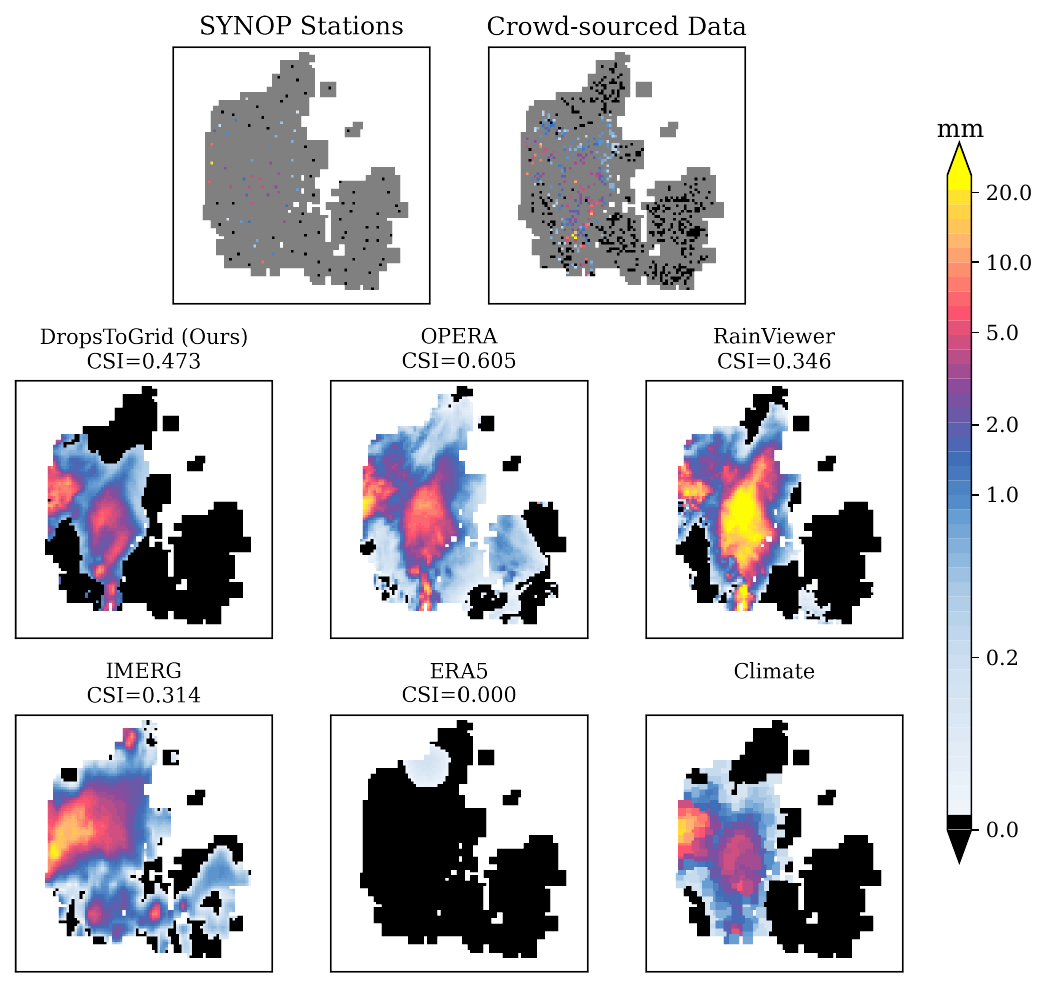}
   \caption{Comparison of rainfall estimators against research-quality SYNOP stations.}
   \label{fig_app:viz_3}
\end{figure*}

\begin{figure*}[p]
  \centering
   \includegraphics[width=0.9\linewidth]{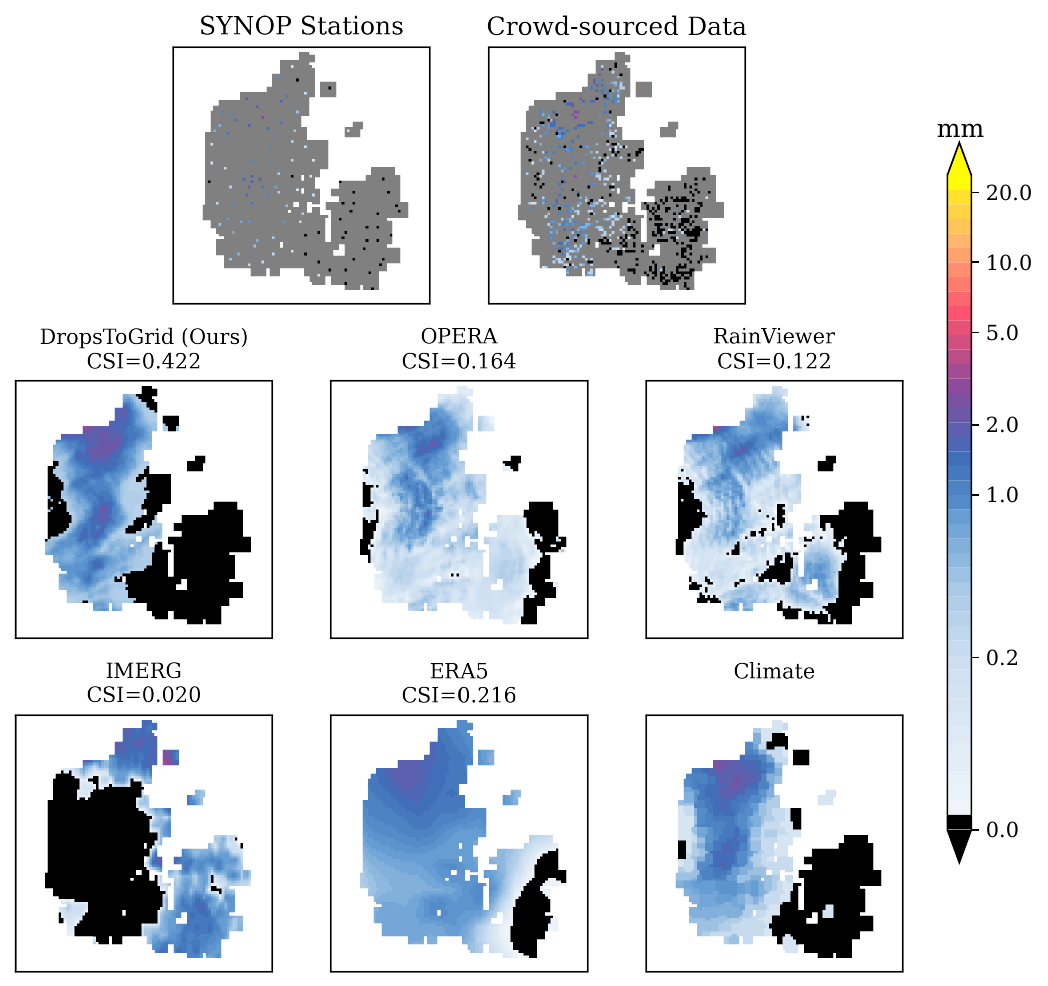}
   \caption{Comparison of rainfall estimators against research-quality SYNOP stations.}
   \label{fig_app:viz_4}
\end{figure*}

Figures~\ref{fig_app:viz_1}-\ref{fig_app:viz_4} show visual comparisons of rainfall estimates from DropsToGrid and several operational baselines. \mbox{DropsToGrid} is derived from crowd-sourced PWS stations and RainViewer radar. 

The baselines include OPERA radar accumulations, RainViewer reflectivity estimates, IMERG satellite retrievals, ERA5 reanalysis, and DMI's gridded \textit{Climate} product. CSI is reported for all products against SYNOP stations, except for Climate, which is excluded since it incorporates the evaluation SYNOP stations in its gridding.

\newpage

\clearpage

%% file: sections/appendix/06_dl_baselines.tex
\clearpage
\section{Deep Learning baselines}
\label{app:dl_baselines}

Beyond the ConvCNP and SwinTNP variants used in the main paper, we evaluate additional baselines. The MM setting uses only station history as input (no radar), while the OOTG setting uses radar and current-time station readings (no history). We further include the translation-equivariant SwinTNP (SwinTNP\_TE) and the approximately translation-equivariant version (SwinTNP\_ATE) from Gridded-TNP~\cite{gridded-tnp} in both settings. Finally, we add an extended ALL ConvCNP model that uses both radar and station history to provide a closer comparison to \mbox{DropsToGrid} in terms of data sources.

\begin{table*}[b]
\centering
\caption{Performance comparison across metrics against deep learning baselines on PWS holdout stations. DropsToGrid achieves superior performance across most metrics. Best results are shown in \textbf{bold}, and second-best in \textit{italics}.}
\label{tab_app:dl_baselines_1}
\begin{tabular}{lllllll}
\toprule
& CSI $\uparrow$ & FSS $\uparrow$ & CRPS $\downarrow$ & FBI $\approx1$ & MAE $\downarrow$ & MSE $\downarrow$ \\
\midrule
\textbf{MM} \\
ConvCNP & 0.484 ± 0.012 & 0.769 ± 0.035 & 0.029 ± 0.000 & 0.849 ± 0.061 & 0.038 ± 0.001 & 0.124 ± 0.003 \\
SwinTNP & 0.407 ± 0.002 & 0.676 ± 0.003 & 0.034 ± 0.000 & 0.766 ± 0.029 & 0.046 ± 0.001 & 0.145 ± 0.002 \\
SwinTNP\_TE & 0.412 ± 0.010 & 0.672 ± 0.021 & 0.034 ± 0.000 & 0.814 ± 0.071 & 0.047 ± 0.001 & 0.147 ± 0.002 \\
SwinTNP\_ATE & 0.418 ± 0.014 & 0.700 ± 0.017 & 0.034 ± 0.001 & 0.772 ± 0.039 & 0.045 ± 0.000 & 0.145 ± 0.003 \\
\midrule
\textbf{OOTG} \\
ConvCNP & \textit{0.486 ± 0.007} & 0.748 ± 0.013 & \textit{0.027 ± 0.000} & 0.752 ± 0.015 & \textbf{0.035 ± 0.000} & \textit{0.115 ± 0.001} \\
SwinTNP & 0.433 ± 0.001 & 0.711 ± 0.006 & 0.032 ± 0.000 & 0.805 ± 0.020 & 0.044 ± 0.000 & 0.140 ± 0.002 \\
SwinTNP\_TE & 0.428 ± 0.002 & 0.684 ± 0.010 & 0.032 ± 0.000 & 0.763 ± 0.012 & 0.043 ± 0.000 & 0.141 ± 0.002 \\
SwinTNP\_ATE & 0.434 ± 0.004 & 0.696 ± 0.010 & 0.032 ± 0.000 & 0.798 ± 0.045 & 0.043 ± 0.001 & 0.142 ± 0.003 \\
\midrule
\textbf{ALL} \\
ConvCNP & 0.485 ± 0.022 & \textit{0.775 ± 0.035} & 0.029 ± 0.000 & \textbf{0.879 ± 0.109} & 0.039 ± 0.001 & 0.146 ± 0.022 \\
DropsToGrid & \textbf{0.532 ± 0.002} & \textbf{0.819 ± 0.000} & \textbf{0.026 ± 0.000} & \textit{0.877 ± 0.019} & \textit{0.035 ± 0.000} & \textbf{0.112 ± 0.001} \\
\bottomrule
\end{tabular}
\end{table*}

\begin{table*}[b]
\centering
\caption{Performance comparison across metrics against deep learning baselines on research-quality SYNOP stations. DropsToGrid achieves superior performance across most metrics. Best results are shown in \textbf{bold}, and second-best in \textit{italics}.}
\label{tab_app:dl_baselines_2}
\begin{tabular}{lllllll}
\toprule
& CSI $\uparrow$ & FSS $\uparrow$ & CRPS $\downarrow$ & FBI $\approx1$ & MAE $\downarrow$ & MSE $\downarrow$ \\
\midrule
\textbf{MM} \\
ConvCNP & 0.507 ± 0.012 & \textit{0.760 ± 0.020} & 0.025 ± 0.000 & 0.863 ± 0.050 & 0.038 ± 0.001 & 0.071 ± 0.001 \\
SwinTNP & 0.424 ± 0.005 & 0.666 ± 0.008 & 0.030 ± 0.000 & 0.797 ± 0.032 & 0.045 ± 0.001 & 0.093 ± 0.002 \\
SwinTNP\_TE & 0.423 ± 0.014 & 0.653 ± 0.027 & 0.030 ± 0.000 & 0.834 ± 0.089 & 0.046 ± 0.001 & 0.094 ± 0.001 \\
SwinTNP\_ATE & 0.437 ± 0.010 & 0.679 ± 0.013 & 0.030 ± 0.001 & 0.780 ± 0.037 & 0.045 ± 0.000 & 0.091 ± 0.003 \\
\midrule
\textbf{OOTG} \\
ConvCNP & \textit{0.522 ± 0.015} & 0.754 ± 0.021 & \textit{0.023 ± 0.000} & 0.823 ± 0.031 & \textbf{0.035 ± 0.000} & \textbf{0.067 ± 0.000} \\
SwinTNP & 0.457 ± 0.005 & 0.700 ± 0.007 & 0.028 ± 0.000 & 0.859 ± 0.035 & 0.043 ± 0.001 & 0.087 ± 0.001 \\
SwinTNP\_TE & 0.439 ± 0.001 & 0.655 ± 0.005 & 0.028 ± 0.000 & 0.790 ± 0.009 & 0.042 ± 0.000 & 0.087 ± 0.001 \\
SwinTNP\_ATE & 0.439 ± 0.007 & 0.665 ± 0.014 & 0.028 ± 0.000 & 0.824 ± 0.045 & 0.043 ± 0.001 & 0.087 ± 0.003 \\
\midrule
\textbf{ALL} \\
ConvCNP & 0.513 ± 0.007 & 0.757 ± 0.006 & 0.025 ± 0.000 & \textit{0.987 ± 0.144} & 0.039 ± 0.001 & 0.090 ± 0.016 \\
DropsToGrid & \textbf{0.551 ± 0.006} & \textbf{0.795 ± 0.007} & \textbf{0.023 ± 0.000} & \textbf{0.995 ± 0.030} & \textit{0.037 ± 0.001} & \textit{0.070 ± 0.005} \\
\bottomrule
\end{tabular}
\end{table*}

\newpage

Across both the PWS holdout and SYNOP stations (Tables~\ref{tab_app:dl_baselines_1}–\ref{tab_app:dl_baselines_2}), DropsToGrid achieves the strongest results on almost all metrics, with higher spatial skill (CSI, FSS), better calibration (CRPS), and lower errors (MAE, MSE). \mbox{ConvCNP} models generally outperform the SwinTNP variants in both MM and OOTG settings, likely because they better capture the local structure of rainfall. Even when provided with both radar and station history, the ALL \mbox{ConvCNP} baseline remains below DropsToGrid, underscoring the benefits of DropsToGrid’s processing and fusion design.

\clearpage

%% file: sections/appendix/07_ablation.tex
\clearpage
\section{Ablation}
\label{app:ablation}

Tables~\ref{tab_app:ablation_1} and \ref{tab_app:ablation_2} summarize the ablation studies conducted to assess the contribution of each component of \mbox{DropsToGrid} on the PWS holdout stations and the research-grade SYNOP stations, respectively. The two primary ablations discussed in the main paper examine (i) replacing the carefully designed fusion bottleneck with a standard convolution that simply stacks all latent source representations in the input (\textit{no\_bottleneck}), and (ii) allowing input stations to also serve as target stations during training (\textit{target\_inputs}), in contrast to DropsToGrid’s strategy of excluding inputs from the prediction targets to avoid direct input-output mapping and mitigate noisy observations. We additionally evaluate variants without station history (\textit{no\_stations}), without radar (\textit{no\_radar}), and with a transformer using standard (non–translation-equivariant) attention (\textit{no\_te}). Regarding the output distribution, beyond the \textit{target\_inputs} ablation, we test a plain gamma distribution that omits the zero-inflation component (\textit{gamma}), as well as a Gaussian output distribution (\textit{gaussian}), to highlight the importance of modeling the highly skewed, zero-dominated nature of rainfall.

\begin{table*}[b]
\centering
\caption{Ablation experiments on PWS holdout stations. DropsToGrid achieves superior performance across most metrics. Best results are shown in \textbf{bold}, and second-best in \textit{italics}.}
\label{tab_app:ablation_1}
\begin{tabular}{lllllll}
\toprule
& CSI $\uparrow$ & FSS $\uparrow$ & CRPS $\downarrow$ & FBI $\approx1$ & MAE $\downarrow$ & MSE $\downarrow$ \\
\midrule
DropsToGrid & \textbf{0.532 ± 0.002} & 0.819 ± 0.000 & \textbf{0.026 ± 0.000} & 0.877 ± 0.019 & \textbf{0.035 ± 0.000} & \textbf{0.112 ± 0.001} \\
\quad no\_bottleneck & 0.520 ± 0.015 & 0.804 ± 0.025 & 0.027 ± 0.000 & 0.868 ± 0.042 & 0.036 ± 0.001 & 0.115 ± 0.005 \\
\quad no\_stations & \textit{0.530 ± 0.004} & 0.825 ± 0.006 & \textit{0.026 ± 0.000} & 0.905 ± 0.006 & 0.036 ± 0.000 & 0.113 ± 0.003 \\
\quad no\_radar & 0.530 ± 0.004 & \textbf{0.829 ± 0.004} & 0.027 ± 0.000 & \textit{0.905 ± 0.011} & 0.036 ± 0.000 & 0.115 ± 0.001 \\
\quad no\_te & 0.530 ± 0.005 & \textit{0.826 ± 0.005} & 0.026 ± 0.000 & 0.871 ± 0.018 & \textit{0.036 ± 0.000} & \textit{0.113 ± 0.001} \\
\quad target\_inputs & 0.514 ± 0.008 & 0.816 ± 0.006 & 0.028 ± 0.001 & \textbf{0.996 ± 0.044} & 0.039 ± 0.001 & 0.149 ± 0.007 \\
\quad gamma & 0.471 ± 0.016 & 0.727 ± 0.024 & 0.032 ± 0.000 & 0.705 ± 0.052 & 0.038 ± 0.001 & 0.118 ± 0.003 \\
\quad gaussian & 0.414 ± 0.006 & 0.637 ± 0.016 & 0.045 ± 0.003 & 0.635 ± 0.027 & 0.041 ± 0.001 & 0.129 ± 0.001 \\
\bottomrule
\end{tabular}
\end{table*}

\begin{table*}[b]
\centering
\caption{Ablation experiments on research-quality SYNOP stations. DropsToGrid achieves superior performance across most metrics. Best results are shown in \textbf{bold}, and second-best in \textit{italics}.}
\label{tab_app:ablation_2}
\begin{tabular}{lllllll}
\toprule
& CSI $\uparrow$ & FSS $\uparrow$ & CRPS $\downarrow$ & FBI $\approx1$ & MAE $\downarrow$ & MSE $\downarrow$ \\
\midrule
DropsToGrid & \textbf{0.551 ± 0.006} & \textit{0.795 ± 0.007} & \textbf{0.023 ± 0.000} & \textbf{0.995 ± 0.030} & \textit{0.037 ± 0.001} & \textbf{0.070 ± 0.005} \\
\quad no\_bottleneck & 0.529 ± 0.023 & 0.770 ± 0.023 & \textit{0.024 ± 0.000} & 0.966 ± 0.058 & \textbf{0.037 ± 0.000} & 0.077 ± 0.006 \\
\quad no\_stations & 0.549 ± 0.003 & 0.794 ± 0.004 & 0.024 ± 0.000 & 1.032 ± 0.019 & 0.037 ± 0.000 & 0.078 ± 0.001 \\
\quad no\_radar & \textit{0.551 ± 0.009} & \textbf{0.796 ± 0.009} & 0.024 ± 0.000 & 1.026 ± 0.025 & 0.037 ± 0.000 & 0.072 ± 0.002 \\
\quad no\_te & 0.542 ± 0.002 & 0.789 ± 0.004 & 0.024 ± 0.000 & \textit{0.989 ± 0.006} & 0.037 ± 0.000 & \textit{0.072 ± 0.003} \\
\quad target\_inputs & 0.505 ± 0.009 & 0.755 ± 0.008 & 0.029 ± 0.001 & 1.198 ± 0.044 & 0.042 ± 0.001 & 0.119 ± 0.014 \\
\quad gamma & 0.485 ± 0.014 & 0.719 ± 0.022 & 0.033 ± 0.000 & 0.769 ± 0.041 & 0.038 ± 0.000 & 0.075 ± 0.003 \\
\quad gaussian & 0.409 ± 0.009 & 0.600 ± 0.016 & 0.046 ± 0.002 & 0.622 ± 0.028 & 0.041 ± 0.001 & 0.081 ± 0.002 \\
\bottomrule
\end{tabular}
\end{table*}

\newpage

As noted in the main paper, the specialized fusion bottleneck is critical, with \textit{no\_bottleneck} showing consistent degradation across all metrics. Although the effects are smaller in magnitude, radar information, station history, and translation-equivariant attention each provide measurable gains, from pixelwise skill (CSI) to calibration of the distribution (CRPS) and overall error (MAE/MSE). Preventing direct input–output mapping is also important, particularly given the noise in PWS observations, as reflected in the significant performance drop against SYNOP stations in \textit{target\_inputs}. Finally, the \textit{gamma} and \textit{gaussian} variants confirm that an appropriate output distribution is essential for rainfall, whose statistics are highly skewed and dominated by zero-rain cases.

\clearpage

%% file: sections/appendix/08_stations.tex
\clearpage
\section{Station analysis}
\label{app:stations}

To assess how varying observational coverage affects \mbox{DropsToGrid}, we study performance under progressively reduced densities of input PWS stations. Starting from all 902 pixels with PWS data, we randomly mask stations in 10\% increments, using nested masks so that each higher-density configuration contains all stations from the previous one. We repeat the experiment with three seeds and average the results, where each seed influences the station masking and model. For each density level, we evaluate the complete test set against SYNOP stations using CRPS and CSI with only the corresponding subset of stations as input.

\begin{figure*}[b]
    \centering
    \includegraphics[width=0.9\linewidth]{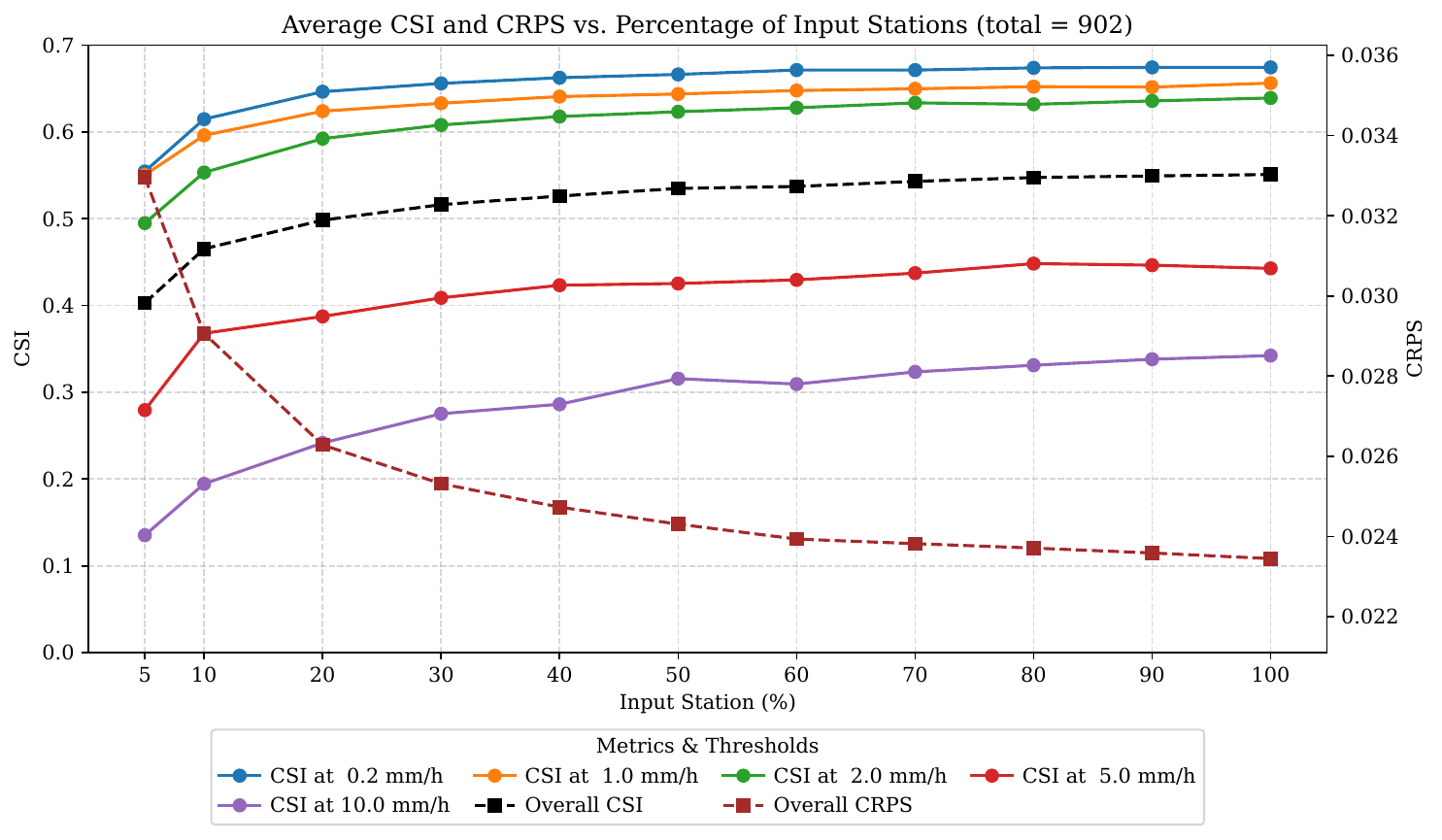}
    \caption{Average performance across station density levels.}
    \label{fig_app:station_plot}
\end{figure*}

\begin{figure}[t]
    \centering
    \includegraphics[width=\linewidth]{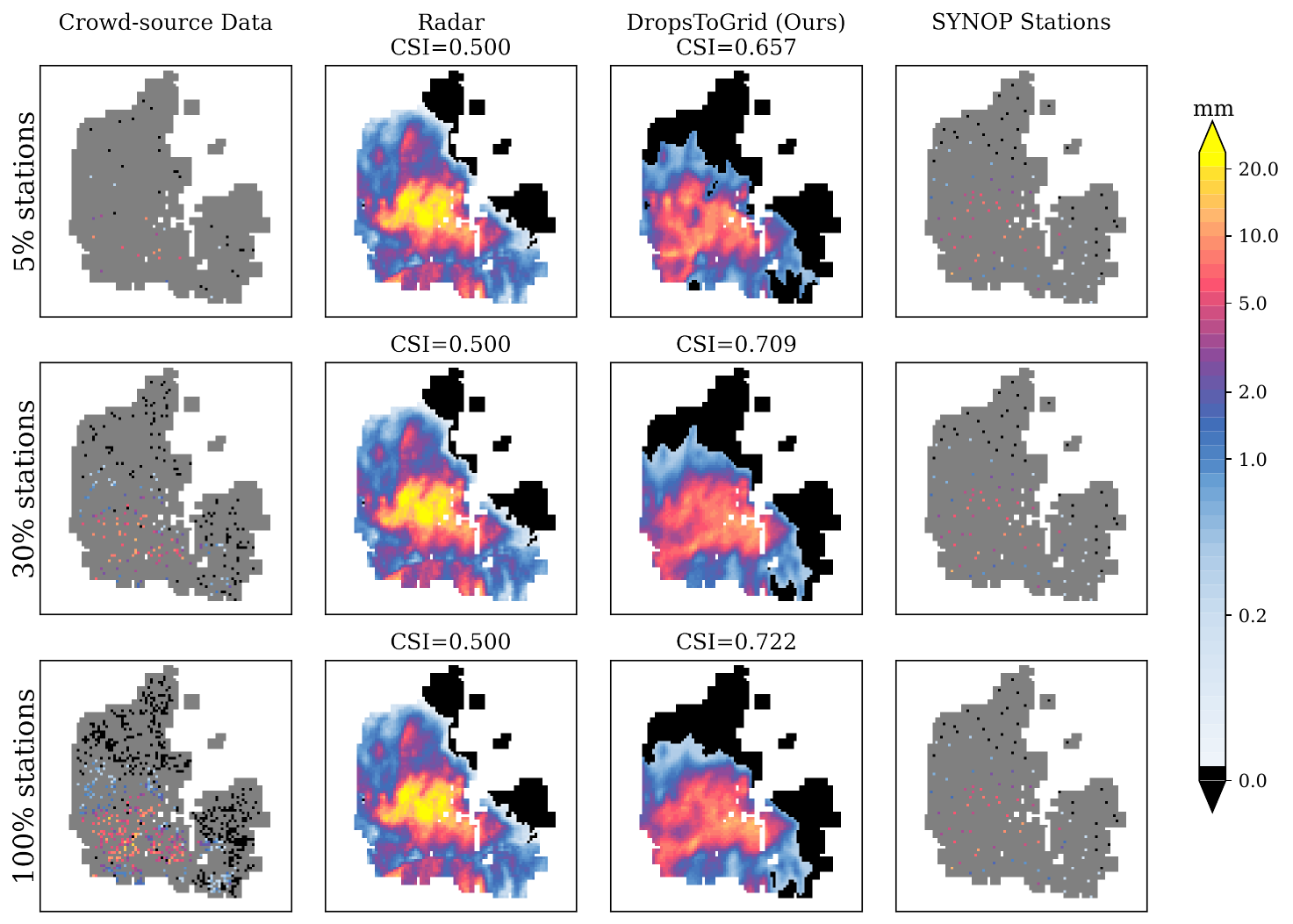}
    \caption{Qualitative example of station masked inputs, radar input, predictions, and targets.}
    \label{fig_app:station_masking}
\end{figure}

DropsToGrid performs best when all stations are available, but remains highly robust even under severe sparsity: at 5\% density, performance remains strong and still surpasses operational estimators such as OPERA (CSI 0.323) by more than 24\% in CSI. This demonstrates the model’s ability to leverage limited point observations while still capturing localized rainfall structure. Figure~\ref{fig_app:station_plot} summarizes the performance trend, and Figure~\ref{fig_app:station_masking} illustrates a qualitative example showing the station inputs (5\%, 30\%, and 100\%), radar observations, the resulting predictions by DropsToGrid, and SYNOP targets, along with the corresponding CSI values. As shown in the case study, with more input stations, the level of detail of DropsToGrid increases.

\clearpage

%% file: sections/appendix/09_eu.tex
\clearpage
\section{Europe-wide densification}
\label{app:europe}

\begin{table*}[b]
  \centering
  \setlength{\tabcolsep}{4.5pt}
  \caption{Performance comparison across regions and station types for all of 2025. \textbf{Note:} PWS and SYNOP stations differ from those in the main paper, going from dense DK to sparser and noiser but wider EU coverage. With worse data quality and access, overall performance is lower, yet, DropsToGrid maintains similar improvements over baselines as in Table~\ref{tab:baselines_stations}, demonstrating robust rainfall estimation.}
  \label{tab:eu_eval}
  \begin{tabular}{lcccccccccccc}
    \toprule
    \multirow{3}{*}{Method}
    & \multicolumn{6}{c}{Denmark}
    & \multicolumn{6}{c}{Europe} \\
    \cmidrule(lr){2-7} \cmidrule(lr){8-13}
    & \multicolumn{3}{c}{PWS holdout}
    & \multicolumn{3}{c}{SYNOP Stations}
    & \multicolumn{3}{c}{PWS holdout}
    & \multicolumn{3}{c}{SYNOP Stations} \\
    \cmidrule(lr){2-4} \cmidrule(lr){5-7}
    \cmidrule(lr){8-10} \cmidrule(lr){11-13}
    & CSI$\uparrow$ & FSS$\uparrow$ & MAE$\downarrow$
    & CSI$\uparrow$ & FSS$\uparrow$ & MAE$\downarrow$
    & CSI$\uparrow$ & FSS$\uparrow$ & MAE$\downarrow$
    & CSI$\uparrow$ & FSS$\uparrow$ & MAE$\downarrow$ \\
    \midrule
    OPERA
    & 0.206 & 0.395 & 0.068 & 0.281 & 0.409 & 0.123 & 0.298 & 0.608 & 0.082 & 0.261 & 0.455 & 0.093 \\
    RainViewer
    & 0.240 & 0.471 & 0.069 & 0.289 & 0.430 & 0.130 & 0.314 & 0.619 & 0.086 & 0.308 & 0.540 & 0.113 \\
    IMERG
    & 0.182 & 0.392 & 0.121 & 0.232 & 0.351 & 0.216 & 0.199 & 0.473 & 0.120 & 0.205 & 0.426 & 0.148 \\
    ERA5
    & 0.157 & 0.350 & 0.103 & 0.197 & 0.299 & 0.161 & 0.144 & 0.321 & 0.119 & 0.171 & 0.324 & 0.137 \\
    Climate
    & 0.388 & 0.713 & 0.057 & - & - & - & - & - & - & - & - & - \\
    \midrule
    ConvCNP\_dk
    & 0.330 & 0.624 & 0.059 & 0.337 & 0.462 & 0.123 & 0.253 & 0.466 & 0.085 & 0.243 & 0.407 & 0.110 \\
    \midrule
    DropsToGrid\_dk
    & \textbf{0.451} & \textbf{0.741} & \textbf{0.046} & \textit{0.465} & \textit{0.606} & \textbf{0.093} & \textit{0.462} & \textit{0.785} & \textit{0.050} & \textit{0.387} & \textit{0.628} & \textbf{0.079} \\
    DropsToGrid\_eu
    & \textit{0.444} & \textit{0.729} & \textbf{0.046} & \textbf{0.472} & \textbf{0.614} & \textit{0.096} & \textbf{0.476} & \textbf{0.802} & \textbf{0.049} & \textbf{0.400} & \textbf{0.646} & \textbf{0.079} \\
    \bottomrule
  \end{tabular}
\end{table*}

We evaluate DropsToGrid across the entirety of Europe for the full year of 2025, a period entirely unseen during training, encompassing a wide range of climatic and topographic conditions. This evaluation requires multiple sources: the WeatherUnderground PWS Network\footnote{\url{https://www.wunderground.com/pws/overview}}, OPERA radar, and GHCNh SYNOP stations\footnote{\url{https://www.ncei.noaa.gov/products/global-historical-climatology-network-hourly}}, all providing at least EU-wide coverage. Table~\ref{tab:eu_eval} presents results in which a model trained exclusively on Danish data (DropsToGrid\_dk) is evaluated across Europe against both operational baselines and a learned baseline (ConvCNP\_dk).

We further demonstrate scalability and improved performance when the same compact, parameter-efficient model is trained on all available EU data (DropsToGrid\_eu). While training still uses patch-based inputs, inference in DropsToGrid is patch-free, achieved through matched TE-Transformer windows, which eliminates border artifacts. EU-wide inference is efficient, taking 0.20s for DropsToGrid versus 0.47s for ConvCNP, the latter being slower due to patch-based processing and 3D convolutions.

Climate baseline is excluded from EU-wide evaluation because it is only available for Denmark and utilizes SYNOP stations. Consequently, Table~\ref{tab:eu_eval} reflects the limitations of global PWS and SYNOP networks, including lower data quality, sparser coverage, and heterogeneous temporal sampling compared to the Denmark-focused dataset (Table~\ref{tab:baselines_stations}), leading to overall performance drops across baselines and learned models. Despite these challenges, DropsToGrid consistently outperforms all baselines. In contrast, ConvCNP trained in Denmark suffers from missing temporal attention and cross-attention mechanisms, limiting its ability to handle longer temporal gaps and heterogeneous spatiotemporal sources.

Overall, DropsToGrid exhibits robust generalization to unseen regions and seasons across Europe in 2025. It learns to suppress noise from PWS temporal observations, rely on radar in sparsely observed areas, and correct biases in radar-only regions by capturing systematic noise patterns. These capabilities allow DropsToGrid to substantially surpass operational baselines, regardless of network quality, coverage, or temporal heterogeneity.

\begin{figure*}[p]
  \centering
   \includegraphics[width=\linewidth]{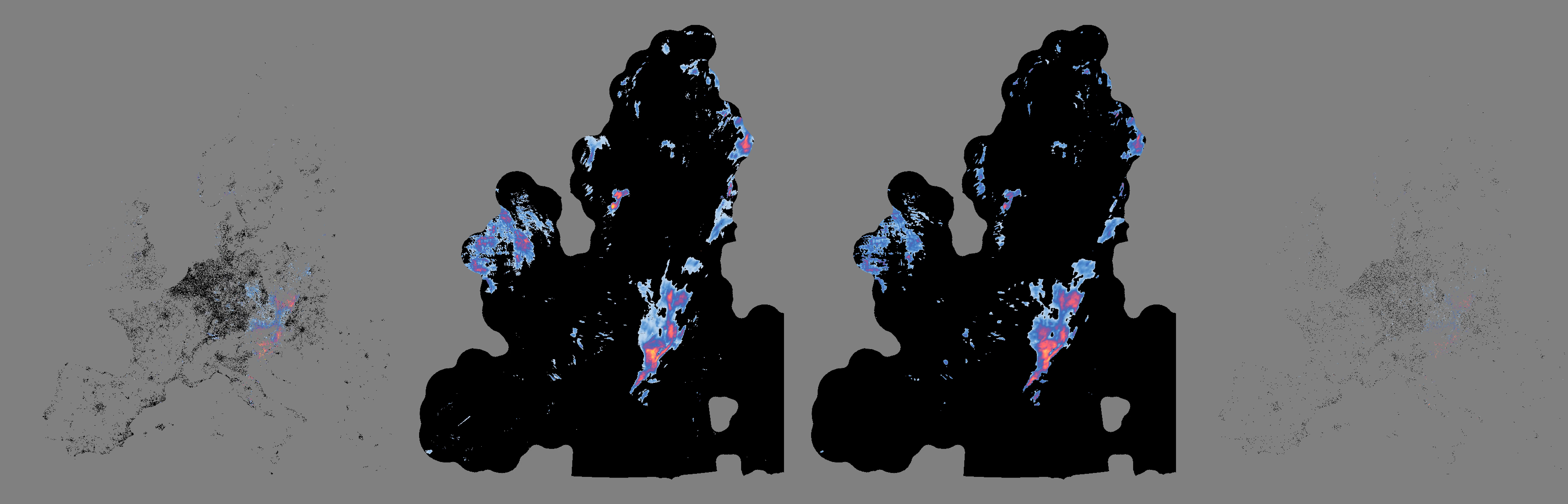}
   \caption{Sample EU densification. From left to right: input PWS stations, input radar, DropsToGrid prediction, and PWS holdout stations.}
   \label{fig_app:eu_viz_1}
\end{figure*}

\begin{figure*}[p]
  \centering
   \includegraphics[width=\linewidth]{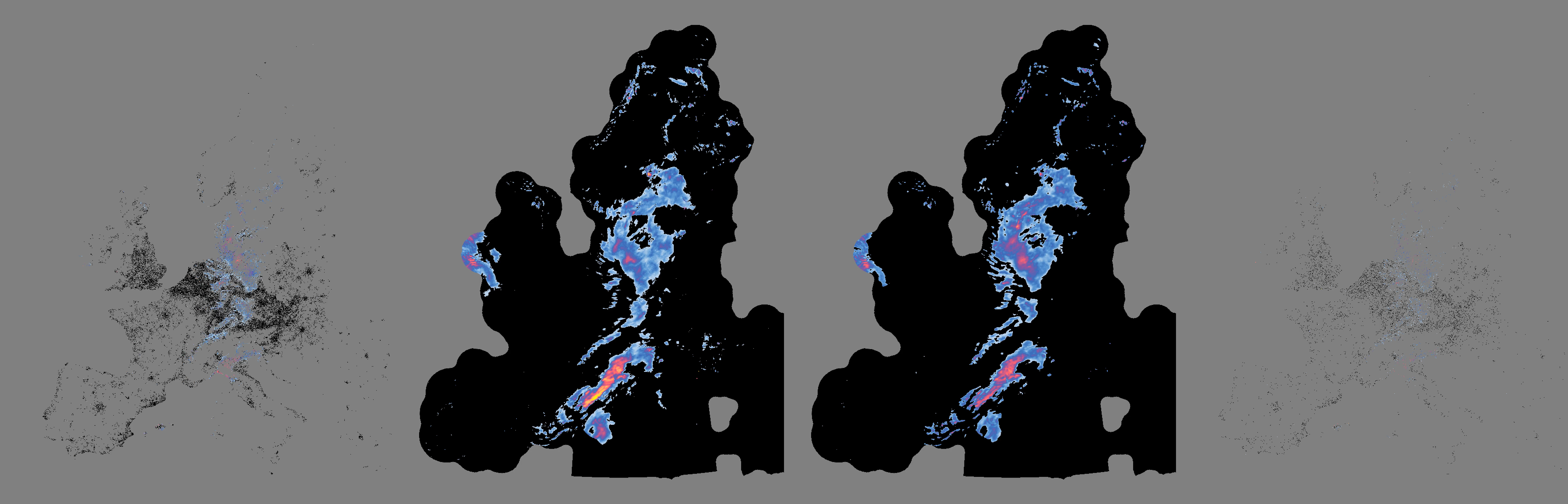}
   \caption{Sample EU densification. From left to right: input PWS stations, input radar, DropsToGrid prediction, and PWS holdout stations.}
   \label{fig_app:eu_viz_2}
\end{figure*}